\documentclass{article}

\PassOptionsToPackage{numbers, compress}{natbib}

 \usepackage[main, final]{neurips_2026}

\usepackage[utf8]{inputenc} 
\usepackage[T1]{fontenc}    
\usepackage{hyperref}       
\usepackage{url}            
\usepackage{booktabs}       
\usepackage{amsfonts}       
\usepackage{nicefrac}       
\usepackage{microtype}      
\usepackage{xcolor}         
\usepackage{enumitem}
\usepackage{amsmath}
\usepackage{algorithm}
\usepackage{algorithmic}
\usepackage{multirow}
\usepackage{graphicx}
\usepackage{subcaption}
\usepackage{tcolorbox}
\usepackage{caption}
\usepackage{subcaption}
\usepackage{hyperref}
\usepackage{wrapfig2}
\usepackage{makecell}

\hypersetup{
    colorlinks=true,
    linkcolor=red,
    citecolor=blue,
    urlcolor=blue
}
\title{Semantics-to-Contact: A Stagewise Framework for Robust Contact-Rich Manipulation}

\author{%
Guanren Qiao$^1$, Ruixiang Ouyang$^1$, Sheng Xu$^1$, Yueci Deng$^{1,2}$, Yunxin Tai$^2$, \\\textbf{Kui Jia}$^{1,2}$, \textbf{Guiliang Liu}$^{1,3}$
\thanks{Correspondence to: Guiliang Liu <liuguiliang@cuhk.edu.cn>.}\\
  $^{1}$The Chinese University of Hong Kong, Shenzhen,\\
  $^{2}$DexForce Co., Ltd,
  $^{3}$Shenzhen Loop Area Institute.\\
  \texttt{liuguiliang@cuhk.edu.cn, guanrenqiao1@link.cuhk.edu.cn} \\
}

\begin{document}

\maketitle

\begin{abstract}
   Contact-rich manipulation requires precise control under complex contact dynamics while remaining robust to diverse visual conditions. However, real-world reinforcement learning often overfits to the visual appearance of training scenes. We propose Semantics-to-Contact (S2C), a stagewise framework for visually robust contact-rich manipulation. The insight is to decompose the problem into two stages: semantic focusing and contact refinement. In the first stage, a vision-language model localizes task-relevant regions across visually diverse scenes, providing coarse spatial guidance and reducing the burden of exploration. In the second stage, a real-world human-in-the-loop residual reinforcement learning policy learns fine-grained contact behaviors within the localized region using force feedback. To further improve visual robustness, we introduce an object-centric visual augmentation strategy that randomizes background appearance while preserving the robot and task-relevant objects. Experiments on six contact-rich real-world manipulation tasks exhibit that S2C consistently improves success rates over strong baselines and maintains reliable performance under significant visual and positional variations. These results demonstrate that stagewise semantic focusing and contact refinement provide a practical path toward visually robust real-world contact-rich manipulation. Our demos can be seen in \url{https://edem-ai.github.io/S2C/}.
\end{abstract}

\section{Introduction}

Contact-rich manipulation is a key capability for enabling robots to operate in real-world applications \cite{luo2025precise}, such as insertion \cite{qiao2026focus}, alignment \cite{zhang2026lira,xu2026reaction}, and pressing \cite{luo2024serl, zhang2026lira}, which often require robots to perform fine-grained adjustments under multimodal force signals and frequently changing contact states. In recent years, reinforcement learning (RL) has demonstrated strong adaptability in such tasks \cite{qiao2025signbot, johannink2019residual, Sutton2018rl, retzlaff2024human}, particularly learning contact policies through human-in-the-loop (HIL) interaction with the real-world environment \cite{luo2024serl, luo2025precise}. However, real-world reinforcement learning (RWRL) still faces a fundamental challenge: \textit{visual observations are highly susceptible to variations in background textures and lighting conditions, making policies learned in one scenario difficult to generalize to new real-world environments}.

Existing methods often rely on domain randomization in simulation to improve visual robustness. By randomly varying background textures, lighting conditions, and environmental parameters, policies can learn representations that are more robust to visual perturbations across a large number of virtual environments \cite{xu2026twinrl, yuan2023rl, wang2025focus}. However, constructing a high-fidelity simulation environment for contact-rich manipulation is challenging \cite{zhang2024bridging}. Contact-rich manipulation tasks often exhibit a large sim-to-real gap \cite{jiang2024transic}. For example, in industrial assembly, tight insertion tasks may involve clearances smaller than 0.1 mm, leading to complex contact dynamics that are difficult to accurately model in simulation. Therefore, a natural question arises: 
\textit{Can we learn visually robust contact policies directly on real robots without relying on complex simulation environments?}

To address the above challenges, we propose \textbf{Semantics-to-Contact (S2C)}, a stagewise framework for visually robust contact-rich manipulation. We argue that a robot is a complex system in which perception, planning, and control are tightly coupled. Therefore, \textbf{visual generalization cannot be achieved by optimizing a single module alone; instead, it requires a stagewise, system-level design} \cite{wang2026learningdeployingfleetscalereinforcement}. The key idea is to decompose the task into two complementary stages: \textit{semantic focusing} and \textit{contact refinement}. In the first stage, a vision-language model (VLM) uses RGB-D observations and language instructions to identify task-relevant regions and guide the robot toward a semantic neighborhood. This stage provides coarse spatial grounding \cite{lu2026worldcoder} across diverse scenes and prevents the RL policy from blindly exploring a large spatial range. In the second stage, human-in-the-loop real-world residual RL learns fine-grained contact behaviors within the localized region using proprioceptive and force feedback. This decomposition is particularly suitable for contact-rich manipulation: once the robot has been guided close to the target, it typically only needs to perform subtle pose adjustments to finish tasks successfully according to local force feedback.

To further enhance visual robustness, we introduce an \textit{Object-centric visual augmentation strategy} during real-world training. Specifically, we extract task-relevant foreground regions from wrist-camera images and apply random perturbations to the background \cite{chen2024focus, wang2025focus}, including changes in color and texture. The augmented images are then directly fed into the RWRL policy for training. In this way, the policy is encouraged to focus on manipulable objects and contact-relevant regions, rather than overfitting to the visual background of a specific experimental setup. At the control level, we adopt \textit{residual RL} \cite{qiao2026focus, johannink2019residual, ankile2025imitation, ankile2025residual} to improve the efficiency of real-world policy learning. Instead of learning a complete action policy from scratch, residual RL learns local corrections on top of an existing policy, thereby reducing the exploration space and improving learning efficiency.

We evaluate our method on six real-world contact-rich manipulation tasks and test its robustness under diverse visual and spatial variations. Experimental results show that, compared with strong baselines, S2C maintains a high success rate under both positional shifts and visual disturbances, while other methods almost fail to complete the task under visual disturbances. In summary, our contributions are threefold: 
\begin{itemize}[leftmargin=*]
    \item We propose \textbf{Semantics-to-Contact (S2C)}, a stagewise real-world reinforcement learning framework that decomposes visually robust contact-rich manipulation into semantic focusing and contact refinement.
    \item We introduce a simulation-free visual robustness strategy that combines VLM-guided semantic localization with an object-centric visual augmentation strategy, enabling RWRL policies to better generalize across background textures and lighting conditions.
    \item We conduct many experiments and provide a number of videos to identify the robustness and scalability boundaries of S2C. The results illustrate that S2C achieves strong performance under most robust settings.
\end{itemize}

\section{Related Work}

\subsection{Reinforcement Learning for Visual Generalization}

Visual reinforcement learning often suffers the challenge of generalization to new environments \cite{cobbe2019quantifying, laskin2020reinforcement, zhou2022domain,yuan2023rl,wang2024effects,wang2026rlscaniqa}. Practically, the previous works can be categorized into two groups: visual processing and RL structure designing. For visual processing, on one hand, \cite{zhang2021learning, fan2022dribo, zang2022simsr} leveraged visual distractions to learn a robust policy from images. For example, VAI \cite{wang2021unsupervised} extracts a universal visual foreground mask to provide an invariant observation to RL. On the other hand, \cite{yarats2020image, laskin2020reinforcement, raileanu2021automatic} focused on image augmentation to improve the generalization performance of the policy. Specifically, FGA3 \cite{lee2025fourier} leveraged adaptive adversarial augmentation while maintained semantic consistency to improve robustness of the agent. For RL structure designing, PAD \cite{hansen2021self} incorporated self-supervision to allow policies continue studying in new environments after deployment. DrQ \cite{yarats2020image} and SVEA \cite{hansen2021stabilizing} leveraged augmented Q and augmented target Q to provide policy generalization performance. However, many of the previous works mainly focused on simulation environments, while our work improves robustness performance in real-world scenarios.

\subsection{Human-in-the-Loop Real-World Reinforcement Learning} 

Recent works successfully integrated human intervention into policy training \cite{liu2026preference,kelly2019hg, retzlaff2024human}, and leveraged RL methods to optimize policy performance, enabling precise and dexterous robot manipulation, such as SERL and HIL-SERL \cite{luo2024serl, luo2025precise}. Leveraging the advantages of human guidance, recent work has greatly improved the robot's performance in dexterous manipulation tasks. HitL integrated human assistance by evaluating the uncertainty of diffusion policies \cite{he2025uncertainty}, while Rac directly learned policy recovery from human corrections for long-horizon tasks \cite{hu2025rac}. Based on a human-in-the-loop RL system, recent works also made progress in fine-tuning VLA models \cite{xu2024rldg, lu2025human, kawaharazuka2025vision}. ConRFT combined behavior cloning and Q-learning to achieve high sample efficiency \cite{chen2025conrft}, while HAPO turned to align the action reference using human interventions, helping VLA models avoid unsatisfactory motions \cite{xia2025human}. However, these works require RL to learn the entire task process, without decoupling the manipulation procedure of each task \cite{xiao2025optical, xiao2025robotic}, and they do not explicitly consider visual robustness.

\section{Problem Formulation}

\paragraph{Goal-Conditioned Partially Observable Markov Decision Process.}
We formulate real-world contact-rich manipulation as an infinite-horizon goal-conditioned partially observable Markov decision process (GC-POMDP), defined by the tuple
$\mathcal{M} = (\mathcal{S}, \mathcal{O}, \mathcal{A}, \mathcal{T}, \mathcal{R}, \mu_0, \gamma, \mathcal{G})$ \cite{Sutton2018rl}.
Here, $\mathcal{S}$ denotes the latent state space of the full robot-environment system, encompassing the states of the robot, objects, and visual background. Since such information is not fully observable in real-world contact-rich manipulation, the agent receives an observation $o_t \in \mathcal{O}$ at each timestep.
Each observation consists of proprioceptive and visual components, i.e.,
$o_t = (o_t^p, o_t^v)$.
The proprioceptive observation $o_t^p$ includes the 6D end-effector pose, linear and angular velocities, and the 6D force-torque measurement. The visual observation $o_t^v$ consists of RGB images captured by wrist-mounted cameras near the end-effector.

The action space $\mathcal{A}$ contains continuous end-effector motion commands and a discrete gripper command. Specifically, each action $a_t \in \mathcal{A}$ consists of a 6D end-effector pose delta and a binary gripper command. The transition function $\mathcal{T}: \mathcal{S} \times \mathcal{A} \rightarrow \Delta(\mathcal{S})$ describes the real-world robot-environment dynamics, and $\mu_0 \in \Delta(\mathcal{S})$ denotes the initial state distribution. The discount factor is denoted by $\gamma \in (0,1)$.

Each task is associated with a goal $g \in \mathcal{G}$, specified by a target contact region. The reward function
$r_t = \mathcal{R}(o_t, a_t, g)$ is a sparse task completion reward. The objective is to learn a goal-conditioned policy
$\pi_\theta(a_t \mid o_t, g)$
that maximizes the expected discounted return $
    J(\pi_\theta)
    =
    \mathbb{E}_{\pi_\theta,\mathcal{T}}
    \left[
    \sum_{t=0}^{\infty}
    \gamma^t r_t
    \right]. $The state-action value function under policy $\pi_\theta$ is defined as
\begin{equation}
Q^{\pi}(o_t, a_t, g)
=
\mathbb{E}_{\pi,\mathcal{T}}
\left[
\sum_{j=t}^{\infty}
\gamma^{j-t} r_j
\mid o_t, a_t, g
\right],
\end{equation}

\paragraph{Real-World Reinforcement Learning with Human Intervention.}
Our real-world RL policy is optimized using Soft Actor-Critic (SAC) \cite{Haarnoja2018sac}. The maximum-entropy objective is
\begin{equation}
\pi^*
=
\arg\max_{\pi}
\mathbb{E}_{\pi,\mathcal{T}}
\left[
\sum_{t=0}^{\infty}
\gamma^t
\left(
r_t
+
\alpha
\mathcal{H}(\pi(\cdot|o_t,g))
\right)
\right],
\end{equation}
where $\alpha$ controls the strength of the regularization of the entropy.

Following human-in-the-loop RWRL \cite{luo2025precise}, a human operator can intervene during training when the robot enters unsafe or unproductive states. We denote the intervention indicator as $l_t \in \{0,1\}$, where $l_t=1$ indicates human control and $l_t=0$ indicates autonomous policy control. The executed action is selected as
\begin{equation}
a_t =
\begin{cases}
a_t^{\mathrm{RL}}, & l_t = 0, \quad a_t^{\mathrm{RL}} \sim \pi_\theta(\cdot|o_t,g), \\
a_t^{\mathrm{H}}, & l_t = 1, \quad a_t^{\mathrm{H}} \sim \pi_{\mathrm{H}}(\cdot|o_t,g).
\end{cases}
\end{equation}

\section{Semantics-to-Contact Stagewise Framework}

\label{sec:method}

\begin{figure}[t]
    \centering
    \includegraphics[width=0.99\textwidth]{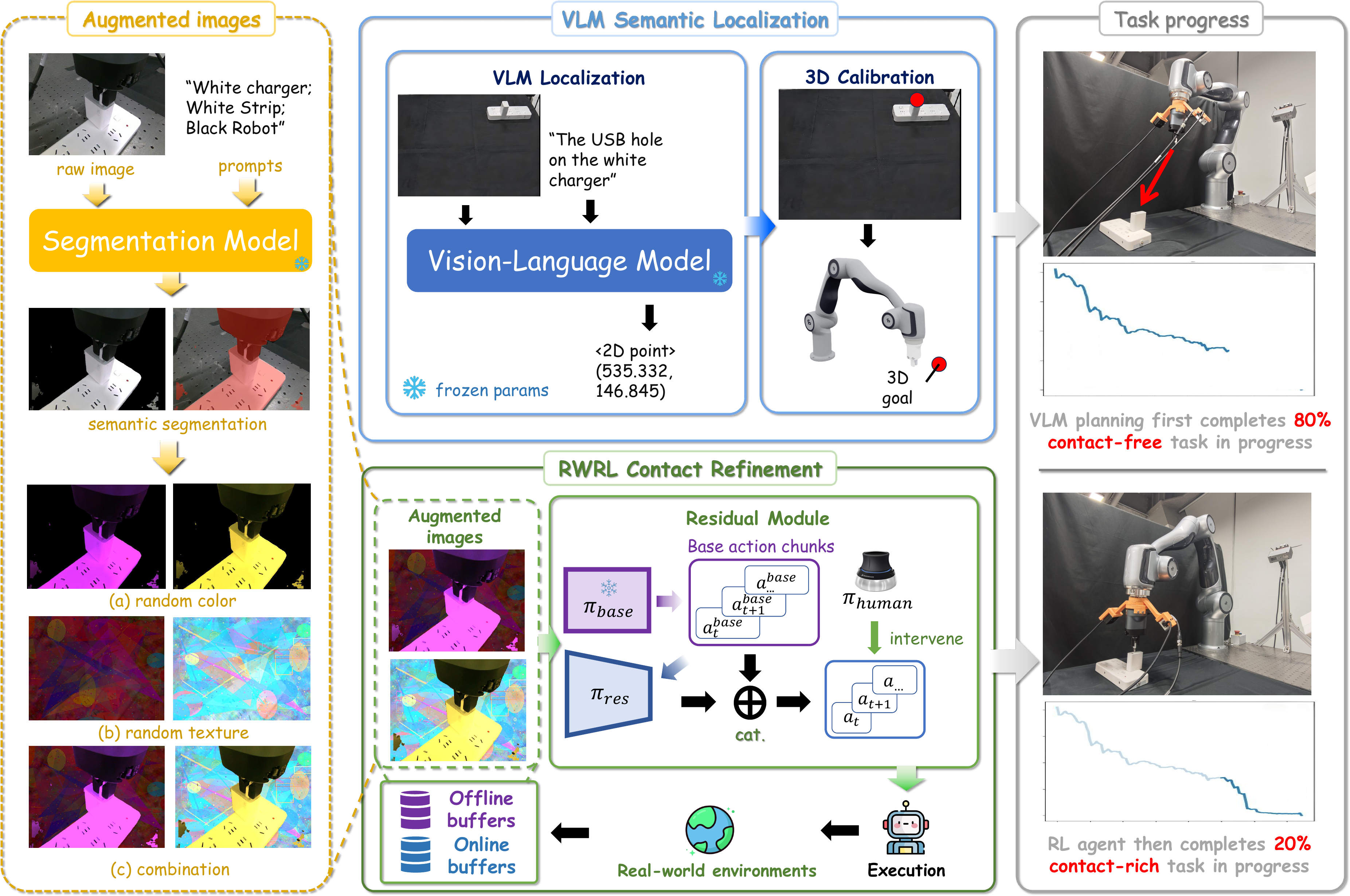}
    \caption{\textbf{S2C Framework}:
    S2C decomposes contact-rich manipulation into two stages: \textit{semantic localization} and \textit{contact refinement}. The semantic focusing stage uses the VLM to predict a task-relevant pre-contact goal $\hat{g}$ and motion planning to move the robot near the desired contact region. The contact refinement stage then uses human-in-the-loop real-world residual RL to learn local force-aware corrections.
    }
    \label{fig:pipeline}
\vspace{-0.2in}
\end{figure}

We present \textbf{Semantics-to-Contact (S2C)}, a stagewise framework for visually generalizable contact-rich manipulation. Figure \ref{fig:pipeline} introduces the pipeline of S2C. First, the robot must understand \textit{where} the task-relevant contact should happen in a visually diverse scene. Second, once it reaches the vicinity of the target, it must determine \textit{how} to make contact under uncertain force interactions. These two problems have different structures: the former requires open-world semantic understanding and spatial grounding, while the latter requires fine-grained closed-loop control with force feedback. S2C explicitly separates these two challenges into two stages: the first stage performs \textit{semantic localization} to identify task-relevant contact regions, while the second stage performs \textit{contact refinement} through real-world residual reinforcement learning to learn fine-grained contact behaviors.

\subsection{Semantic Localization: VLM-Guided Active Perception}
\label{subsec:vlm_brain}

The first stage addresses a core difficulty in RWRL: the agent does not know \textit{where} to explore in the visual space. Without prior semantic information, the RL policy would need to wander randomly over large 3D workspace areas, which is sample-inefficient and often unsafe on a real robot. We leverage a VLM to close this semantic gap.

\paragraph{VLM Guidance.} At the beginning of each episode, the robot receives a language instruction $l$ (e.g., ``Insert into the USB port'') and a third-person view RGB-D observation $(P_t^{side}, D_t)$, where $P_t^{side}$ is the third-view RGB image and $D_t$ is the corresponding depth map. The VLM processes $P_t$ together with $l$ and outputs a prediction point. In practice, we utilize an SOTA VLM MolmoPoint 8B \cite{clark2026molmopoint} with a carefully designed few-shot prompt that asks for the 2D point relevant to $l$. This model is trained to point by directly selecting which visual token contains the target object
and then refining that location by generating additional tokens. Compared to the open-source GPT-5 \cite{singh2025openai, xiao2026yang} or Gemini-Robotics-ER-1.5 \cite{team2025gemini} API, MolmoPoint 8B has stronger image understanding capabilities, enabling it to infer the locations of affordances in an image, while also offering much faster inference speed (2s) than the GPT-5 API ($\geq10s$). In scenarios similar to USB insertion, it can understand where the insertion hole is located in the image, thus meeting our requirements. Using the depth map $D_t$, we compute the 3D position of the interactive object in camera coordinates:
\begin{align}
\mathbf{c}_l^t
=
D_t(u_l^t, v_l^t)
\mathbf{K}^{-1}
[u, v, 1]^{\top},
\quad
\mathbf{c}_l^t \in \mathbb{R}^3 .
\end{align}
where $K$ is the intrinsic matrix of the camera and $u, v$ represent the positions of the points marked by the VLM in the image. We obtain the hand-eye calibration matrix $T_{\text{cam}}^{\text{base}} \in SE(3)$, which denotes the homogeneous transformation matrix from the camera coordinate to the robot base coordinate through a 6-well plate. We fix the position of the camera from a third-person perspective and perform calibration using the eye-to-hand method. The target contact point in the robot base frame is $T_{\text{cam}}^{\text{base}} \cdot c_l$. The robot then executes a motion-planning primitive to reach $p_{\text{target}}^{\text{base}}$. Formally, we define the semantic focus operator:
\begin{equation}
\Phi_{\text{VLM}}(P_t^{side}, l) = p_{\text{target}}^{\text{base}}=T_{\text{cam}}^{\text{base}} \cdot c_l \in SE(3).
\end{equation}
This operator reduces the effective state space for the subsequent RL from the entire workspace to a small local volume.

\paragraph{\bf Object-centric Visual Augmentation.}
\label{subsec:yoloe_aug}

To improve visual robustness without relying on simulation, we introduce an object-centric visual augmentation module \cite{wang2025yoloe} during RWRL training. We choose YOLOE as the segment model because it is lightweight enough for online preprocessing and supports promptable object detection and segmentation. This allows us to specify task-relevant foreground regions using language prompts or visual prompts, without training a task-specific segmentation model from scratch. We choose prompt guidance over visual because the wrist-mounted perspective moves along with the robotic arm, so the reference images provided would not offer effective guidance. Given two wrist-camera images $\{P_{t,k}^{\mathrm{wrist}}\}_{k=1}^{2}$ and a set of prompts $\mathcal{L}=\{l_1,l_2,\ldots,l_N\}$ describing the robot arm, gripper, and manipulated objects, YOLOE predicts a foreground mask for each view:
\begin{equation}
    M_{t,k}
    =
    f_{\mathrm{YOLOE}}
    \left(
    P_{t,k}^{\mathrm{wrist}},
    \mathcal{L}
    \right),
    \quad k\in\{1,2\},
    \label{eq:yoloe_mask}
\end{equation}
where $M_{t,k}\in\{0,1\}^{H\times W}$ is a binary foreground mask. Pixels with $M_{t,k}=1$ correspond to regions that should be preserved, including the robot arm, gripper, manipulated object, and contact-relevant structures, while pixels with $M_{t,k}=0$ correspond to background regions.

For each wrist-camera image, we sample a random background texture $B_{t,k}$ from an augmentation distribution $q(B)$ and construct a background-randomized image:
\begin{equation}
    \bar{P}_{t,k}^{\mathrm{wrist}}
    =
    M_{t,k}\odot P_{t,k}^{\mathrm{wrist}}
    +
    (1-M_{t,k})\odot B_{t,k},
    \label{eq:bg_aug_image}
\end{equation}
where $\odot$ denotes element-wise multiplication. The augmentation distribution includes random colors, textures, and visual distractors. This operation randomizes the appearance of task-irrelevant background while preserving the geometric and semantic structure of the foreground.

In addition to background randomization, we also perturb the segmented foreground to prevent the policy from overfitting to specific object colors or lighting conditions. Let $\mathcal{A}_{\mathrm{fg}}(\cdot)$ denote a foreground augmentation operator that applies random color jitter and brightness changes \cite{chen2024focus}. The final augmented wrist-camera image is:
\begin{equation}
    \tilde{P}_{t,k}^{\mathrm{wrist}}
    =
    M_{t,k}\odot
    \mathcal{A}_{\mathrm{fg}}
    \left(
    P_{t,k}^{\mathrm{wrist}}
    \right)
    +
    (1-M_{t,k})\odot B_{t,k}.
    \label{eq:fg_bg_aug_image}
\end{equation}

The resulting augmented observation is:
$
    \tilde{o}_t
    =
    \left(
    o_t^p,
    \tilde{P}_{t,1}^{\mathrm{wrist}},
    \tilde{P}_{t,2}^{\mathrm{wrist}}
    \right)
    \label{eq:aug_obs}
$. Appendix \ref{appendix: Augmentation} supplements other augmentation details. Unlike simulation-based domain randomization, it does not require modeling and 3D assets. The physical interaction remains real, while task-irrelevant visual factors are randomized at the image level.

\subsection{Contact Refinement: Human-in-the-loop Residual Reinforcement Learning}
\label{subsec:rl_cerebellum}

Once the robot reaches the pre-contact region, the remaining challenge is to establish local contact. This stage is dominated by small pose errors and force feedback. These factors are difficult to model accurately, but can be learned through real-world interaction. We therefore utilize human-in-the-loop real-world residual RL to finish contact refinement.

\paragraph{Base Policy Pre-training.}
We first collect a set of demonstration trajectories:
$
    \mathcal{D}_{\mathrm{expert}}
    =
    \left\{
    (\tilde{o}_t, a_t^{\mathrm{expert}})
    \right\},
    \label{eq:demo_dataset}
$
where $a_t^{\mathrm{expert}}$ denotes the demonstrated action. We use the Action-Chunking Transformer (ACT) \cite{zhao2023learning, shen2026mftformer} as our base policy, which is trained by behavior cloning:
\begin{equation}
    \theta_{\mathrm{base}}^{*}
    =
    \arg\min_{\theta}
    \mathbb{E}_{(\tilde{o}_t,a_t^{\mathrm{expert}})\sim \mathcal{D}_{\mathrm{expert}}}
    \left[
    \left\|
    \pi_{\theta}^{\mathrm{base}}(\tilde{o}_t)
    -
    a_t^{\mathrm{expert}}
    \right\|_2^2
    \right].
    \label{eq:act_bc}
\end{equation}
The ACT policy provides a reasonable initial behavior and reduces the amount of random exploration required during RWRL. However, because behavior cloning is limited by demonstration coverage, it may still fail under visual shifts, pose perturbations, and contact uncertainty.

\paragraph{Residual Reinforcement Learning.}
Instead of replacing the ACT policy, we use it as a base policy and learn a residual correction:
\begin{equation}
    a_t^{RL}
    =
    a_t^{\mathrm{base}}
    +
    \Delta a_t,
    \qquad
    a_t^{\mathrm{base}}
    =
    \pi_\theta^{\mathrm{base}}(\tilde{o}_t),
    \qquad
    \Delta a_t
    \sim
    \pi_{\theta}^{\mathrm{res}}(\cdot \mid \tilde{o}_t, a_{base}, g).
    \label{eq:residual_act}
\end{equation}
Here, $\pi_{\theta}^{\mathrm{res}}$ is the residual policy optimized by RWRL. The residual action $\Delta a_t$ only needs to correct errors of the ACT policy. This significantly reduces the effective exploration space compared with learning the full action policy from scratch.

We update our optimization objective with:
\begin{equation}
    \pi_{\theta}^{\mathrm{res}*}
    =
    \arg\max_{\pi_{\theta}^{\mathrm{res}}}
    \mathbb{E}
    \left[
    \sum_{t=0}^{\infty}
    \gamma^t
    \left(
    r_t
    +
    \alpha
    \mathcal{H}
    \left(
    \pi_{\theta}^{\mathrm{res}}(\cdot|\tilde{o}_t, a_{base}, g)
    \right)
    \right)
    \right],
    \label{eq:residual_sac}
\end{equation}

\paragraph{Human-in-the-loop Residual Reinforcement Learning System.} During training, we deploy the VLM as the server side, where images captured by a third-person camera are used by the VLM to annotate object positions. The base policy and RL act as the client side and drives the robotic arm through impedance control, while the views from two wrist cameras serve as the visual inputs for RL. Meanwhile, during the human-in-the-loop process, we also adopt a client-server architecture within the RWRL. The client is responsible for interacting with the environment, either by allowing the robot to act autonomously according to the base policy or by enabling human intervention. The server is responsible for updating the actor and critic components of the RL policy. We use ResNet-10 as a pretrained encoder and follow HIL-SERL \cite{luo2025precise} to transform the state representation from the robot base frame into an end-effector-centered coordinate frame. Algorithm \ref{alg:sftc_algorithm} presents the overall procedure of the S2C framework.

\begin{algorithm}[t!]
\caption{Semantics-to-Contact}
\label{alg:sftc_algorithm}

    \begin{algorithmic}[1]
        \STATE Randomly initialize Critic $Q_i$ (set targets $\tilde{Q}_i = Q_i$) for $i = 1,2$ and Actor $\pi_\theta^{res}$. Choose discount $\gamma$, temperature $\eta$, batch size $N$, delay ratio $c$, $\mathcal{D}_{online}$
        and $\mathcal{D}_{offline}$. Load base policy $\pi_\theta^{base}$.
        \WHILE{True}
            
            \FOR{$t = 0$ to $T$}
                \STATE Get focused region with $\Phi_{\text{VLM}}(P_t^{side}, l)$
                \STATE Get augmented images with Eq. \ref{eq:fg_bg_aug_image} $
    \tilde{o}_t
    =
    \left(
    o_t^p,
    \tilde{P}_{t,1}^{\mathrm{wrist}},
    \tilde{P}_{t,2}^{\mathrm{wrist}}
    \right),
    \label{eq:aug_obs}
$
                \STATE Take residual action $a^{res}_t \sim \pi_\theta^{res}(\cdot|\tilde{o}_t, a^{base}_t)$ and base action $a^{base}_t \sim \pi_\theta^{base}(\cdot|\tilde{o}_t)$
                \STATE  $a^{RL}_t$ = $a^{base}_t$ + $a^{res}_t$
                or Take human intervention $a^{H}_t \sim \pi_{H}(\cdot|o_t)$
                \STATE $\mathcal{D}_{\mathrm{online}}
    \leftarrow
    \mathcal{D}_{\mathrm{online}}
    \cup
    \{(\tilde{o}_t,a_t^{ RL/H},r_t,\tilde{o}_{t+1})\}.$
                \STATE $\mathcal{D}_{\mathrm{offline}}
    \leftarrow
    \mathcal{D}_{\mathrm{offline}}
    \cup
    \{(\tilde{o}_t,a_t^{\mathrm{H}},r_t,\tilde{o}_{t+1})\}.$
                \FOR{$1$ to $c$}
                    \STATE Sample batch $b_{online}$ of $\frac{N}{2}$ from $\mathcal{D}_{online}$
                    and batch $b_{offline}$ of $\frac{N}{2}$ from $\mathcal{D}_{offline}$
                    \STATE $y_t = r_t + \gamma \min_{i \in \{1,2\}} \tilde{Q}_{i}(\tilde{o}_{t+1}, a^{RL}_{t+1}) + \gamma \eta \log \pi_\theta^{res}(\cdot|\tilde{o}_{t+1}, a_t^{base})$ with $\mathbf{b}$
                    \STATE Update Q networks with $\mathcal{L}^{critic} = \frac{1}{N} \sum (y_t - \mathop{min}\limits_{i=\{1, 2\}}Q_i(\tilde{o}_t,a^{RL}_t))^2$
                \ENDFOR
                \STATE With $\mathbf{b}$, update $\pi_\theta^{res}$ maximizing objective: $\mathbb{E}\left[ \min\limits_{i=\{1, 2\}} Q_i(\tilde{o}_t, a^{RL}_t)- \eta \log \pi_\theta^{res}(\cdot|\tilde{o}_t, a_t^{base}) \right]$
            \ENDFOR
        \ENDWHILE
    \end{algorithmic}

\end{algorithm}
\vspace{-0.1in}

\section{Experiments}
\vspace{-0.1in}
\begin{figure*}[htbp]
    \centering

    \begin{subfigure}[t]{0.24\textwidth}
        \centering
        \includegraphics[width=\linewidth]{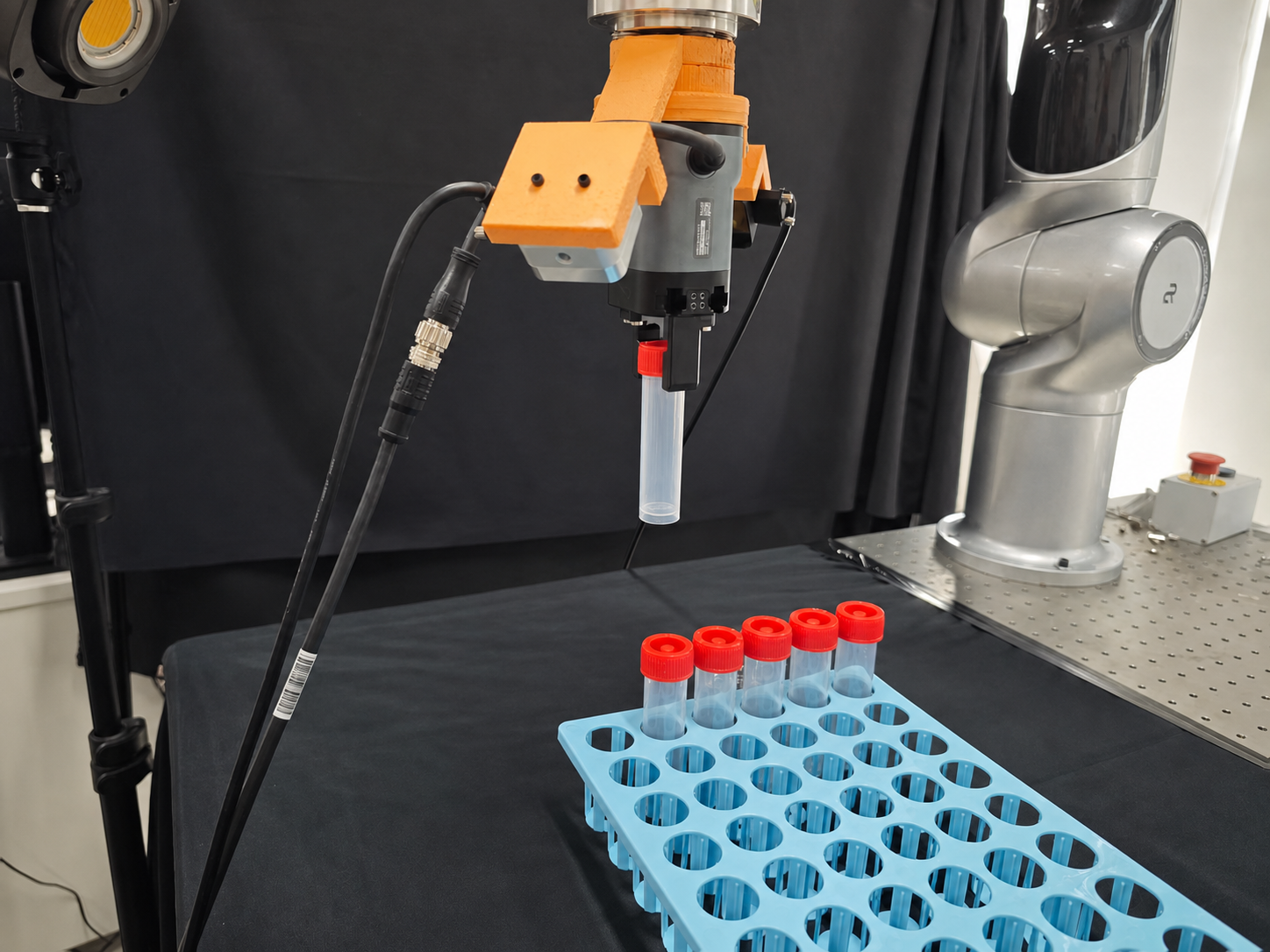}
        \caption{Original}
    \end{subfigure}
    \begin{subfigure}[t]{0.24\textwidth}
        \centering
        \includegraphics[width=\linewidth]{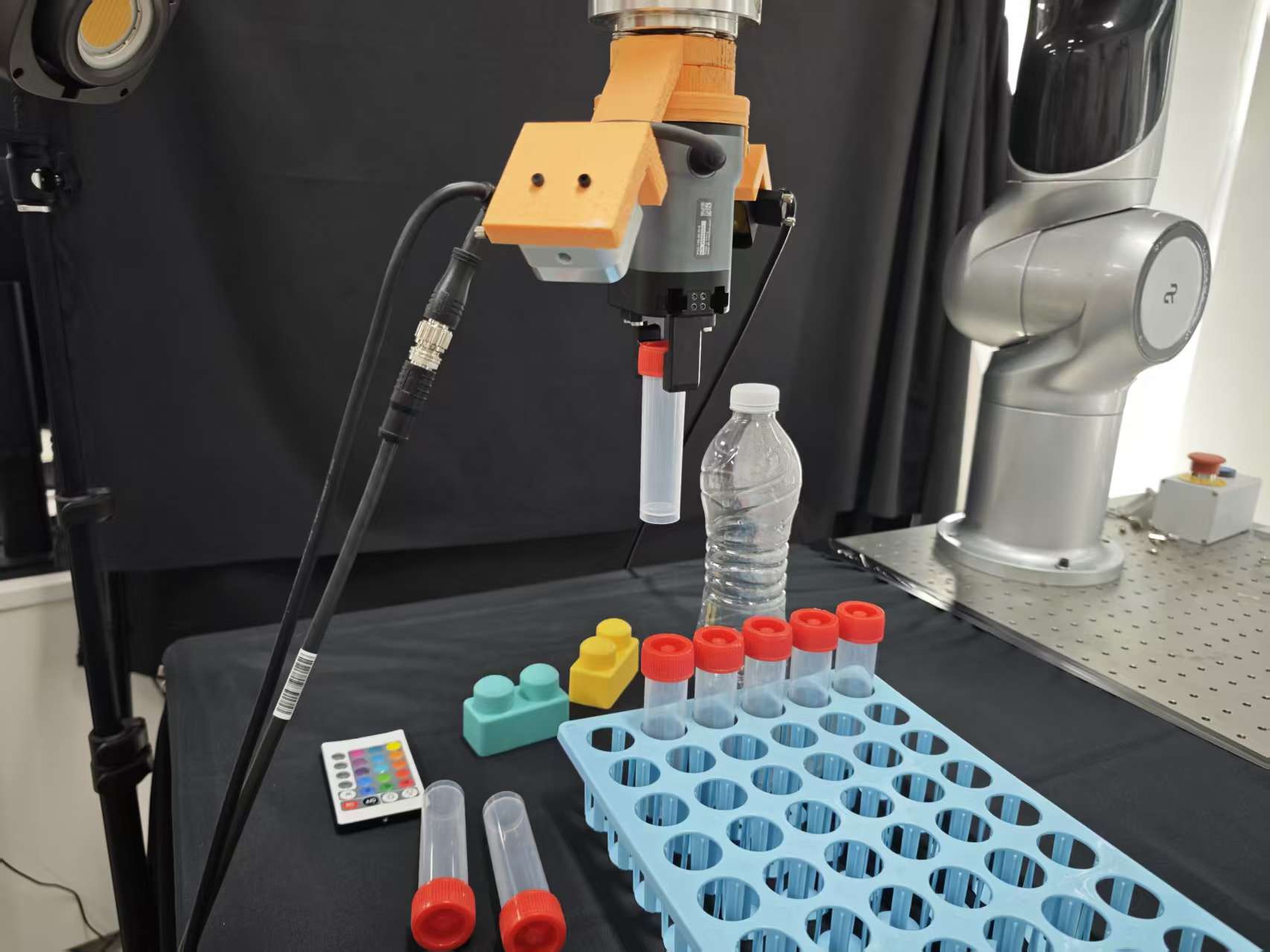}
        \caption{Placing clutter}
    \end{subfigure}
    \begin{subfigure}[t]{0.24\textwidth}
        \centering
        \includegraphics[width=\linewidth]{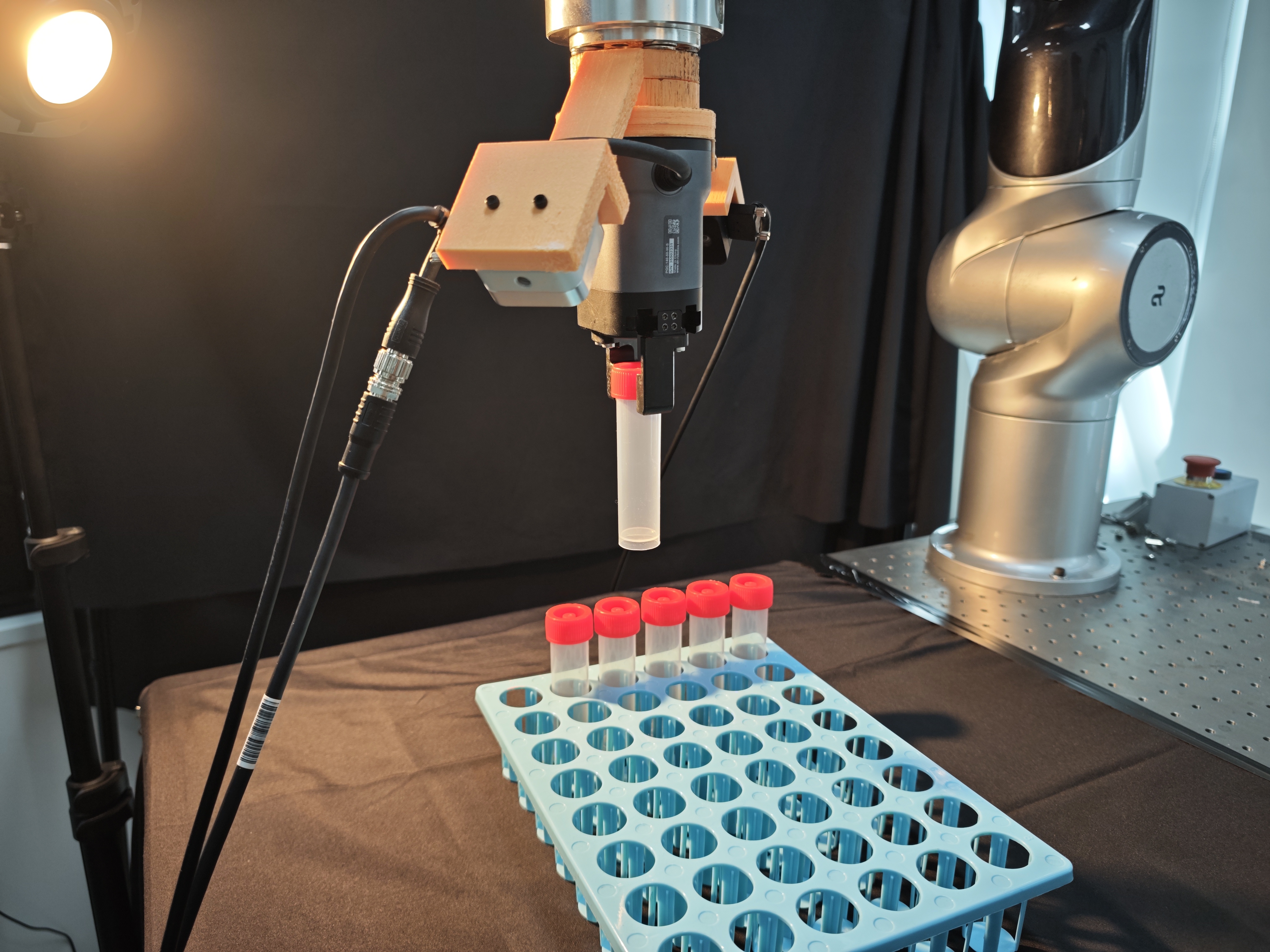}
        \caption{Strong sun light}
    \end{subfigure}
    \begin{subfigure}[t]{0.24\textwidth}
        \centering
        \includegraphics[width=\linewidth]{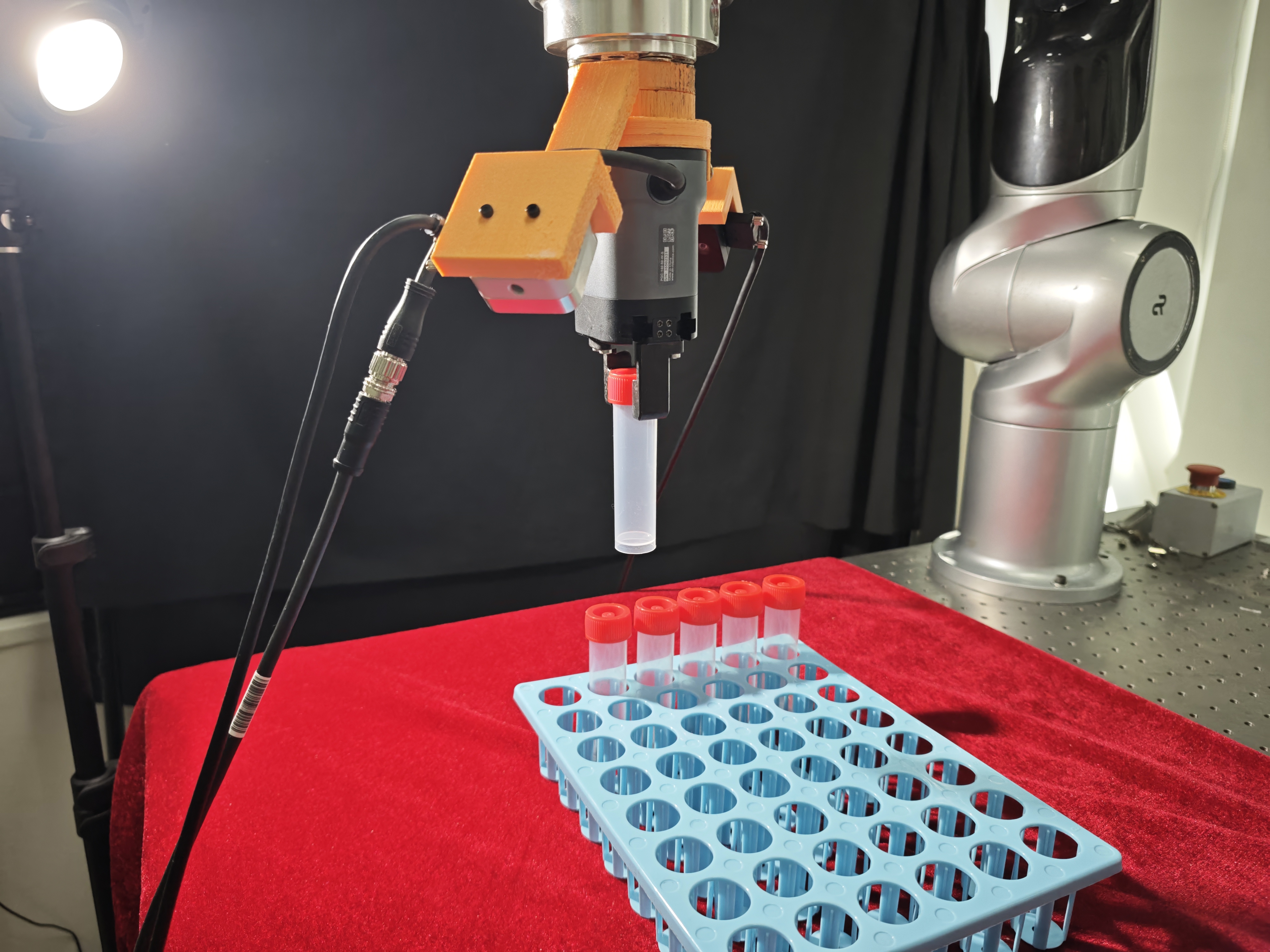}
        \caption{Flashing white light+ Red tablecloth}
    \end{subfigure}

    \vspace{0.5em}

    \begin{subfigure}[t]{0.24\textwidth}
        \centering
        \includegraphics[width=\linewidth]{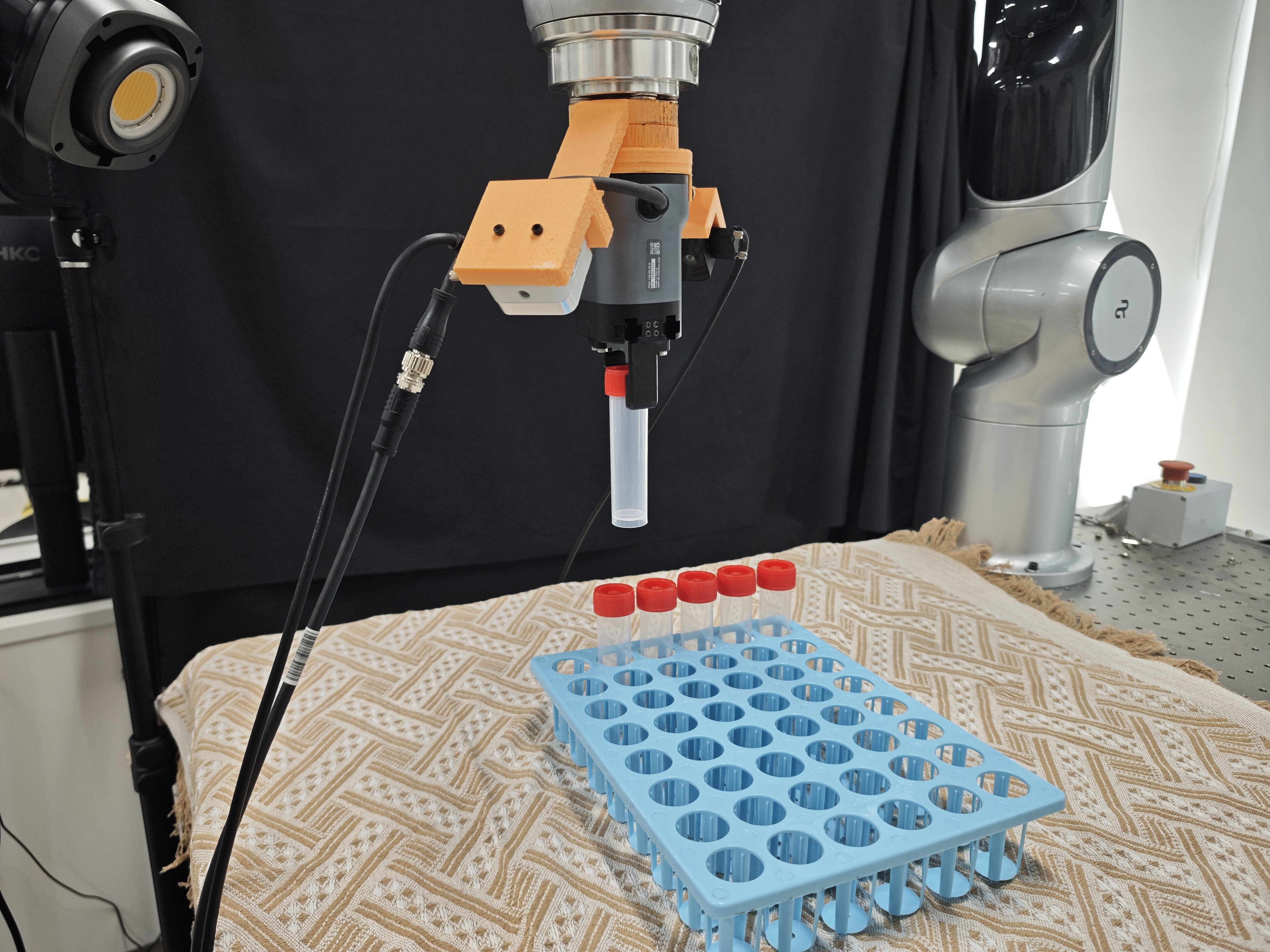}
        \caption{Patterned tablecloth}
    \end{subfigure}
    \begin{subfigure}[t]{0.24\textwidth}
        \centering
        \includegraphics[width=\linewidth]{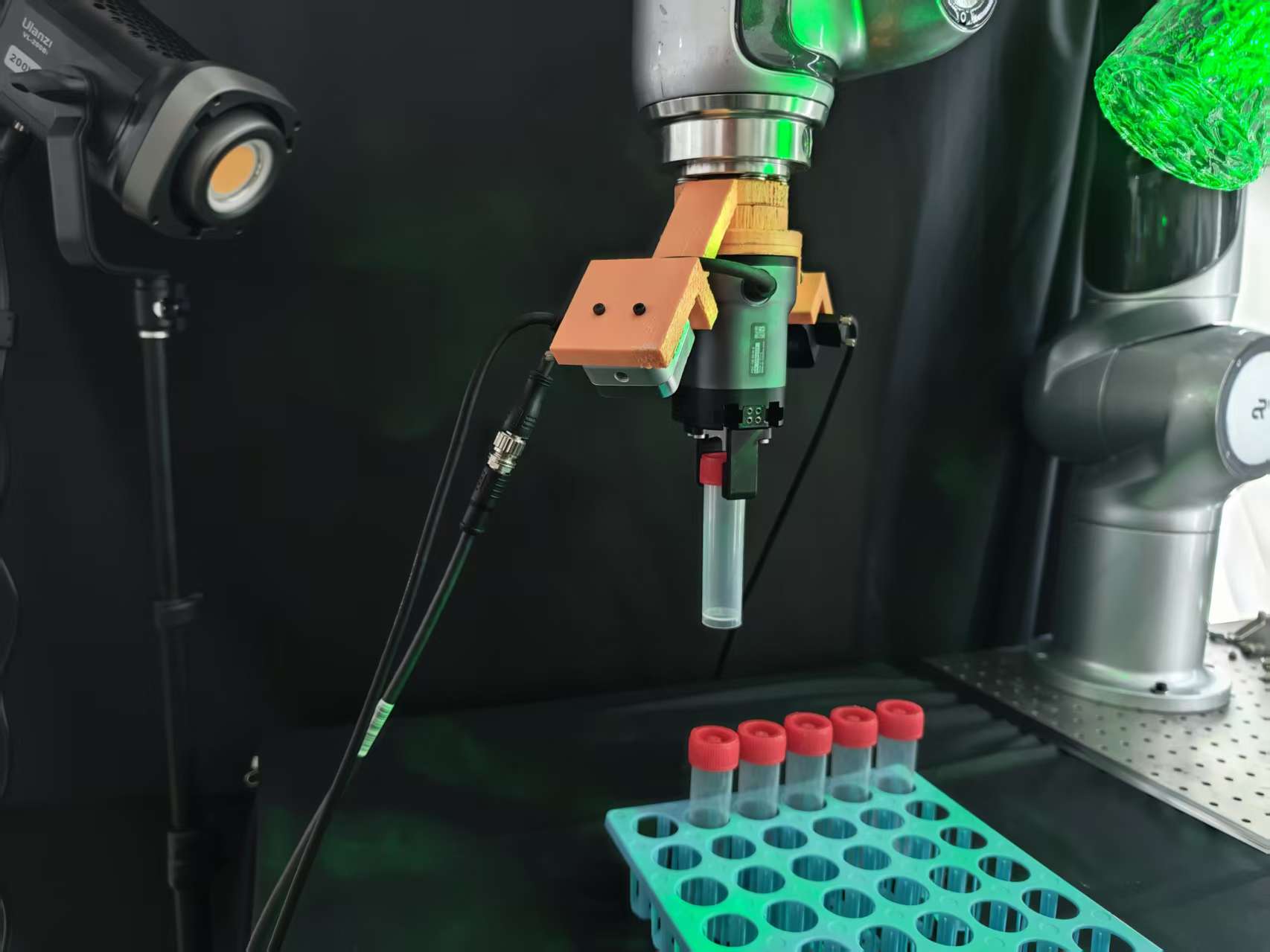}
        \caption{Color light}
    \end{subfigure}
    \begin{subfigure}[t]{0.24\textwidth}
        \centering
        \includegraphics[width=\linewidth]{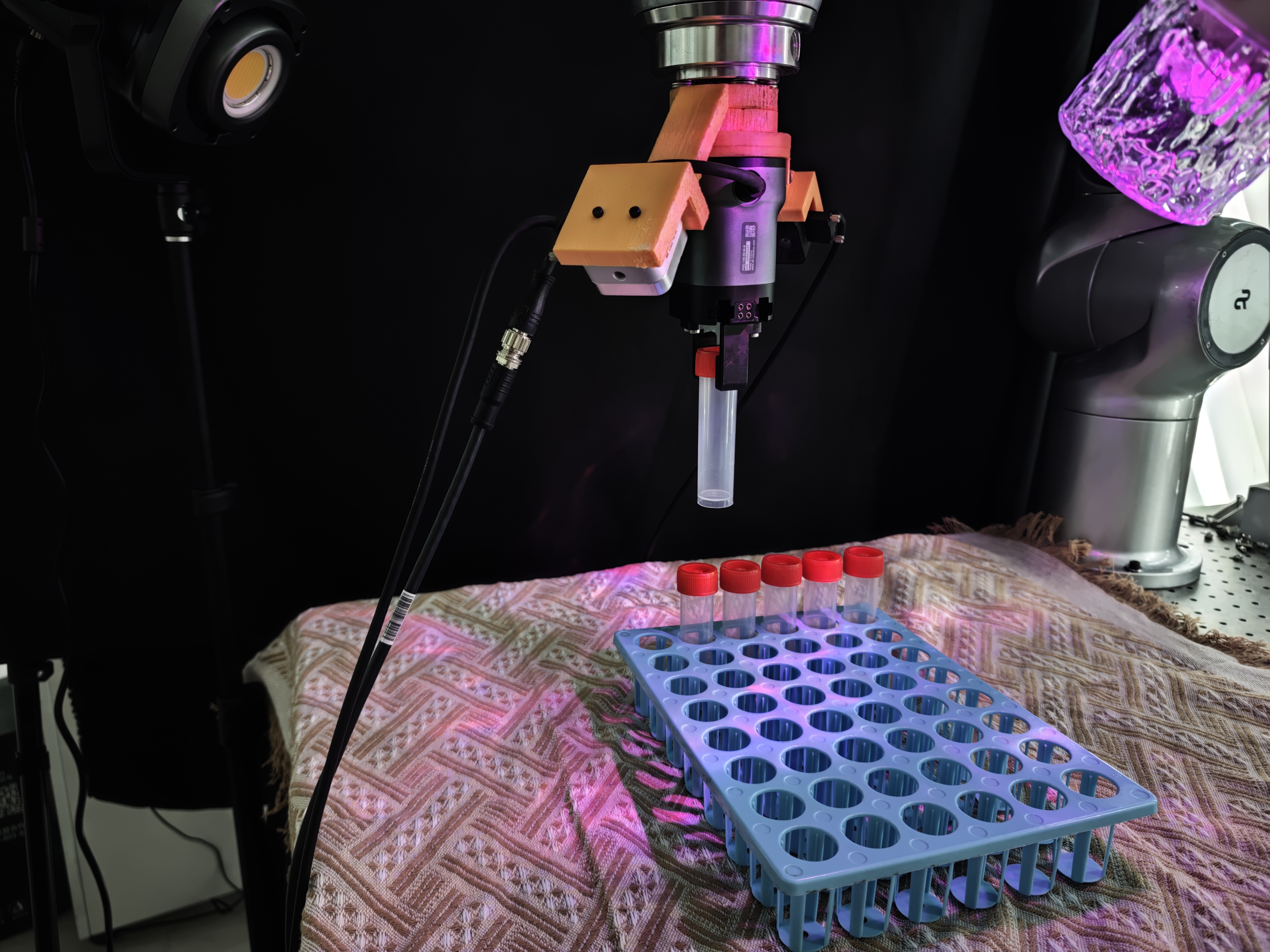}
        \caption{Color light + Patterned tablecloth}
    \end{subfigure}
    \begin{subfigure}[t]{0.24\textwidth}
        \centering
        \includegraphics[width=\linewidth]{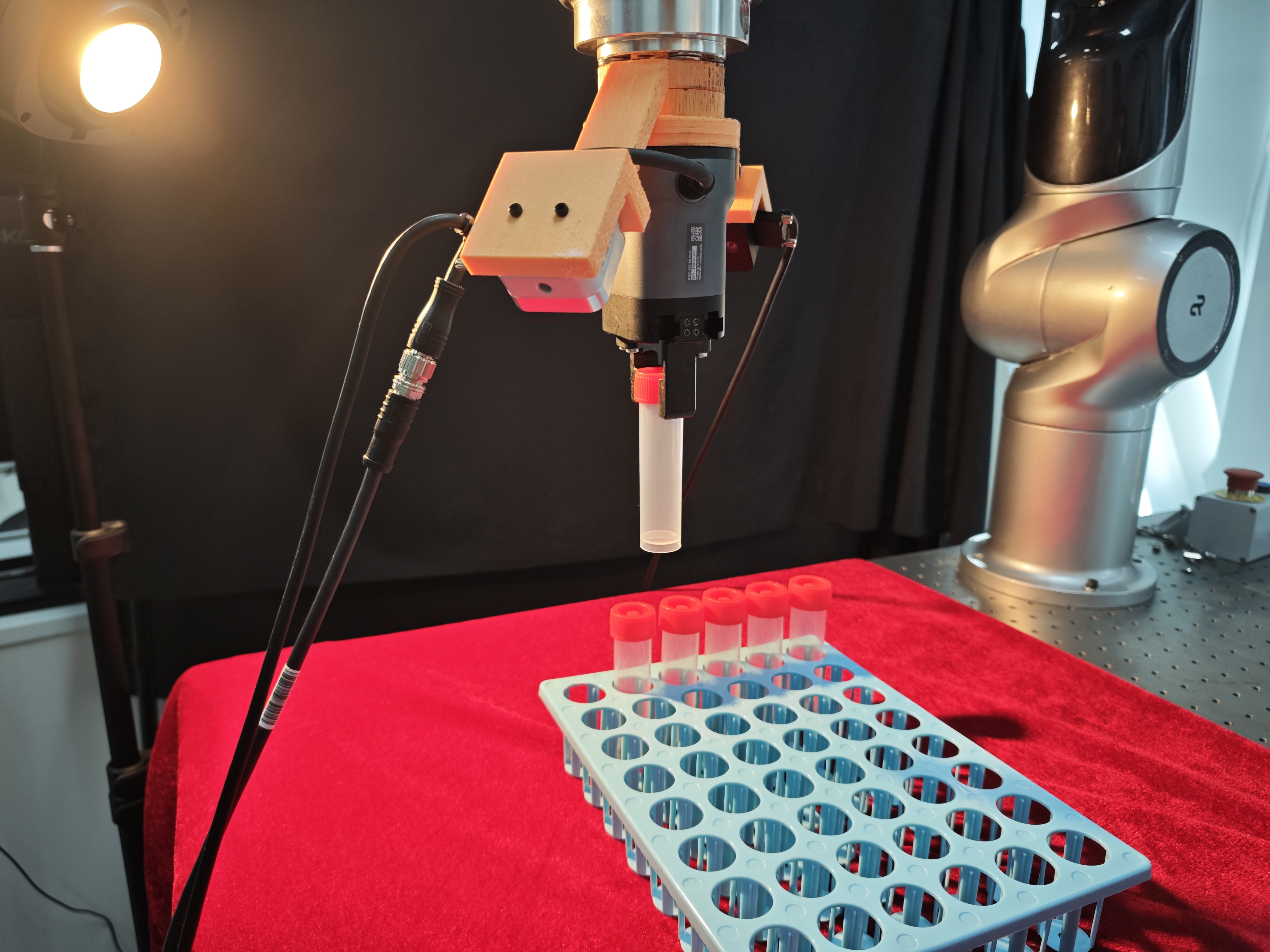}
        \caption{Strong sun light + Red tablecloth}
    \end{subfigure}

    \caption{Different visual robustness settings.}
    \label{fig:eight_images}
    \vspace{-0.2in}
\end{figure*}

\noindent{\bf Experiment Settings.} To conduct a comprehensive evaluation, we quantify the performance of S2C in six contact-rich manipulation tasks from the following perspectives:
1) {\it Robustness}: How well does S2C perform with diverse contact-rich manipulation tasks under position and visual changes?
2) {\it Complementarity}: Is each component of S2C necessary?
3) {\it Scalability}: Where are the scalability boundaries of S2C?

We use the Agile Diana force-controlled robotic arm, which has 7 degrees of freedom per arm and employs impedance control. However, interaction with the arm is still achieved by sending end-effector pose commands. For teleoperation, we use a 3D SpaceMouse to send incremental end-effector changes to the robotic arm during both human-in-the-loop operation and data collection. The cameras we employ are RealSense D405 and Kingfisher depth cameras. VLM prompts can be checked in Appendix \ref{sec: vlm_instructions}.

\noindent{\bf Task Description.} We design six contact-rich atomic skill manipulation tasks, including:
(1) \textit{Charger grasp-insertion}: Grasp a charger and put it into a power strip. (2) \textit{USB insertion (easy)}: Insert a USB plug into a fixed charger. (3) \textit{USB insertion (hard)}: Insert a USB plug into a charger wherever it is located. (4) \textit{Test tube insertion (easy)}: Insert the test tube into the fixed well slot. (5) \textit{Test tube insertion (medium)}: Insert a row of six test tubes in sequence. (6) \textit{Test tube insertion (hard)}: Insert into the empty slots among a row of test tubes, where the empty positions are random. Please refer to Appendix \ref{appendix: task_setup} for other task description and training details. Appendix \ref{appendix: running_setting} displays the running setting for each baseline. The entire experiment is conducted in a natural industrial environment without any black cloth occlusion. Figure \ref{fig:eight_images} presents our different visual robustness settings.

\noindent{\bf Metrics.} In our experiments, we adopt two key metrics: (1) \textit{Success Rate (SR)}: The ratio of successful task completions within the allotted time to the total number of test trials. (2) \textit{Training Time (TT)}: Across all experimental settings, we record the online training time (min).

\noindent{\bf Baselines.} Our method is compared with five other approaches: (1) \textit{ACT} \cite{zhao2023learning}: Action Chunking with Transformers, a method that leverages transformer architecture to generate action sequences in chunks for robotic manipulation.
(2) \textit{SERL} \cite{luo2024serl}: A RWRL framework without human-in-the-loop. 
(3) \textit{Octo-small} \cite{team2024octo}: A large transformer-based policy trained on 800k trajectories from the Open X-Embodiment dataset.
(4) \textit{HIL-SERL}: The first human-in-the-loop vision-based RWRL system.
(5) \textit{HIL-ConRFT} \cite{chen2025conrft}: A RWRL reinforced fine-tuning approach for VLA
model.

\begin{table*}[t]
\centering
\caption{Training time and success rate under training setting. Each setting is tested 20 times. We implement HIL-SERL and HIL-ConRFT following the experimental settings described in their original papers; therefore, they did not use the same VLM-based pre-contact localization and motion-planning initialization as S2C. To ensure consistency and comparison of the success rate and training time, all methods are compared within the same workspace and start from the same initial region in each experiment. We also use the same evaluation protocol and success criterion for all methods, such that a trial is counted as successful only when the object is fully inserted into the target slot. For training time, we report the learner-side actor–critic optimization time because real-robot rollout collection and actor–critic optimization run as separate processes.}
\label{tab:black_pos_generalization}
\resizebox{\textwidth}{!}{
\begin{tabular}{lccccccccccccc}
\toprule
\multirow{2}{*}{Method} 
& \multicolumn{2}{c}{Charger} 
& \multicolumn{2}{c}{USB-Easy} 
& \multicolumn{2}{c}{USB-Hard} 
& \multicolumn{2}{c}{Tube-Easy} 
& \multicolumn{2}{c}{Tube-Medium} 
& \multicolumn{2}{c}{Tube-Hard}
& \multirow{2}{*}{Avg. SR} \\
\cmidrule(lr){2-3} \cmidrule(lr){4-5} \cmidrule(lr){6-7}
\cmidrule(lr){8-9} \cmidrule(lr){10-11} \cmidrule(lr){12-13}
& TT & SR & TT & SR & TT & SR & TT & SR & TT & SR & TT & SR & \\
\midrule
ACT          & -- & 0/20  & --  & 1/20  & --  & 0/20  & --  & 2/20  & --  & 0/20  & --  & 0/20  & 2.50\% \\
SERL         & 120 & 0/20 & 120 & 0/20  & 120 & 0/20  & 120 & 0/20  & 120 & 0/20  & 120 & 0/20  & 0.00\% \\
Octo-small   & -- & 0/20  & --  & 2/20  & --  & 0/20  & --  & 5/20  & --  & 0/20  & --  & 0/20  & 5.83\% \\
HIL-SERL     & \textbf{30} & 19/20 & 60  & \textbf{20/20} & 120 & 18/20 & 30  & \textbf{20/20} & 120 & 16/20 & 120 & 16/20 & \textbf{90.83\%} \\
HIL-ConRFT   & \textbf{30} & 18/20 & 40  & \textbf{20/20} & \textbf{60}  & 19/20 & \textbf{20}  & \textbf{20/20} & 100  & 16/20 & 100  & \textbf{17/20} & \textbf{91.76\%} \\
S2C          & \textbf{30} & \textbf{20/20} & \textbf{30}  & \textbf{20/20} & \textbf{60}  & \textbf{20/20} & \textbf{20}  & \textbf{20/20} & \textbf{90}  & \textbf{17/20} & \textbf{90}  & 16/20 & \textbf{94.17\%} \\
\bottomrule
\end{tabular}
}
\vspace{-0.1in}
\end{table*}

\begin{table*}[t]
\centering
\caption{Success rate under visual and positional robustness.  Only HIL-based RWRL methods and S2C are evaluated. Each setting is evaluated 20 times.}
\label{tab:visual_pos_generalization}
\resizebox{\textwidth}{!}{
\begin{tabular}{lccccccc}
\toprule
Method 
& Charger 
& USB-Easy 
& USB-Hard 
& Tube-Easy 
& Tube-Medium 
& Tube-Hard
& Avg. SR \\
\midrule

\multicolumn{8}{c}{\textbf{Patterned tablecloth + Position}} \\
\midrule
HIL-SERL   & 0/20  & 0/20  & 0/20  & 0/20  & 0/20  & 0/20  & 0.00\% \\
HIL-ConRFT & 0/20  & 0/20  & 0/20  & 0/20  & 0/20  & 0/20  & 0.00\% \\
S2C        & \textbf{18/20} & \textbf{15/20} & \textbf{11/20} & \textbf{13/20} & \textbf{14/20} & \textbf{12/20} & \textbf{69.17\%} \\

\midrule
\multicolumn{8}{c}{\textbf{Strong sun light + Position}} \\
\midrule
HIL-SERL   & 0/20  & 0/20  & 0/20  & 0/20 & 0/20 & 0/20 & 0.00\% \\
HIL-ConRFT & 0/20  & 0/20  & 0/20  & 0/20 & 0/20 & 0/20 & 0.00\% \\
S2C        & \textbf{20/20} & \textbf{17/20} & \textbf{15/20} & \textbf{20/20} & \textbf{16/20} & \textbf{13/20} & \textbf{84.17\%} \\

\midrule
\multicolumn{8}{c}{\textbf{Color light + Position}} \\
\midrule
HIL-SERL   & 0/20  & 0/20  & 0/20  & 0/20  & 0/20  & 0/20  & 0.00\% \\
HIL-ConRFT & 0/20  & 0/20  & 0/20  & 0/20  & 0/20  & 0/20  & 0.00\% \\
S2C        & \textbf{18/20} & \textbf{18/20} & \textbf{19/20} & \textbf{16/20} & \textbf{12/20} & \textbf{12/20} & \textbf{79.17\%} \\

\midrule
\multicolumn{8}{c}{\textbf{Flashing white light + Red tablecloth + Position}} \\
\midrule
HIL-SERL   & 0/20  & 0/20  & 0/20  & 6/20 & 2/20 & 3/20  & 9.17\% \\
HIL-ConRFT & 0/20  & 0/20  & 0/20  & 8/20 & 4/20 & 4/20 & 13.33\% \\
S2C        & \textbf{18/20} & \textbf{19/20} & \textbf{17/20} & \textbf{17/20} & \textbf{15/20} & \textbf{13/20} & \textbf{82.50\%} \\


\bottomrule
\end{tabular}
}
\vspace{-0.2in}
\end{table*}

\subsection{Evaluate the impact of position and visual changes in contact-rich tasks.
}

We first evaluate whether the learned policies can generalize to spatial shifts and visual appearance changes in real-world contact-rich manipulation. Table~\ref{tab:black_pos_generalization} reports training time and success rate under the black-tablecloth setting with positional generalization. Imitation learning or pretrained policies, including ACT and Octo-small, achieve only limited success, indicating that behavior cloning are insufficient for precise contact-rich execution under shifted target positions. SERL without human intervention fails on all tasks. In contrast, HIL-SERL and HIL-ConRFT achieve average success rates of 90.83\% and 91.76\%, respectively, while S2C achieves 94.17\%. The residual design and VLM guidance also enable S2C to complete the tasks with only a short training time.

Table~\ref{tab:visual_pos_generalization} further evaluates the robustness of the learned policies under combined visual and positional changes. Although HIL-SERL and HIL-ConRFT perform well under the training black-tablecloth setting, their performance drops significantly when the background texture or lighting condition changes. This suggests that these policies are highly sensitive to task-irrelevant visual cues and tend to overfit to the appearance of the training scene. In contrast, S2C maintains substantially higher success rates in various visual disturbances, achieving an overall average success rate of 78.75\% in all visual and positional robustness settings. S2C remains highly effective under red tablecloth and strong lighting conditions, with average success rates of 82.5\% and 84.17\%, respectively. Performance decreases under more challenging tablecloth and color light perturbations, but S2C still achieves success rates of 69.17 \% and 79.17 \% in these settings. These results indicate that VLM guidance and Object-centric visual augmentation improve robustness to visual background and lighting variations by enforcing the policy to focus on the contact-relevant regions.

\subsection{Verify the complementarity of each core component of
S2C.}

We further conduct ablation experiments to verify the complementarity of the core components of S2C. Table~\ref{tab:ablation} evaluates different variants under a challenging setting that combines strong sunlight, a red tablecloth, and positional perturbations. HIL-SERL and HIL-ConRFT also use standard image augmentations in their original papers, including random cropping, random shifting. Removing either HIL or the contact module results in complete failure across all tasks. The failure of the w/o HIL variant indicates that real-world reinforcement learning remains extremely difficult without human corrections during unsafe or unproductive exploration. Similarly, the 0\% success rate of w/o Contact demonstrates that semantic localization alone is insufficient for contact-rich manipulation: although the semantic module can guide the robot toward the target region, it cannot resolve fine-grained contact uncertainties or perform force-sensitive corrections during insertion.

\begin{wraptable}{r}{0.5\textwidth}
  \centering
  \caption{Fine-grained ablation of the semantic components. The ablation is conducted on Charger Insertion and USB Insertion (hard). ``w/o VLM'' removes VLM-guided motion planning, whereas ``w/o OC-Aug'' removes object-centric augmentation. Each setting is evaluated 20 times.}
  \label{tab:ablation}
  \footnotesize
  \begin{tabular}{lccc}
    \toprule
    Method & Charger & USB (Hard) & Avg. SR \\
    \midrule
    \multicolumn{4}{c}{\textbf{Strong sun light + Red tablecloth + Position}} \\
    \midrule
    w/o HIL     & 0/20  & 0/20  & 0.00\%  \\
    w/o VLM     & 10/20 & 12/20 & 55.00\% \\
    w/o OC-Aug  & 0/20  & 0/20  & 0.00\%  \\
    w/o Contact & 0/20  & 0/20  & 0.00\%  \\
    S2C         & \textbf{16/20} & \textbf{14/20} & \textbf{75.00\%} \\
    \midrule
    \multicolumn{4}{c}{\textbf{Placing clutter around objects}} \\
    \midrule
    w/o HIL     & 0/20  & 0/20  & 0.00\%  \\
    w/o VLM     & 13/20 & 10/20 & 57.50\% \\
    w/o OC-Aug  & 2/20  & 1/20  & 7.50\%  \\
    w/o Contact & 0/20  & 0/20  & 0.00\%  \\
    S2C         & \textbf{16/20} & \textbf{15/20} & \textbf{77.50\%} \\
    \bottomrule
  \end{tabular}
\end{wraptable}

To further evaluate robustness under extreme visual and spatial disturbances, we have performed a more fine-grained ablation study, as shown in Table \ref{tab:ablation}. Under strong sunlight, a red tablecloth, and position variation, “w/o OC-Aug” fails on both tasks, whereas full S2C achieves 16/20 on Charger Insertion and 14/20 on USB Insertion. Similarly, in cluttered scenes, “w/o OC-Aug” achieves only 2/20 and 1/20, compared with 16/20 and 15/20 for full S2C. Because VLM guidance is retained in this ablation, these large performance gaps directly show that reliable semantic localization alone is insufficient to prevent the local policy from overfitting to task-irrelevant visual cues. nevertheless, the VLM guidance provides an auxiliary role. With object-centric enhancement retained, “w/ no VLM achieves 10/20 and 12/20 under combined lighting, background, and position perturbation, and 13/20 and 10/20 in cluttered scenes. Restoring VLM guidance improves these results to 16/20 and 14/20 in the former setting and to 16/20 and 15/20 in the latter. The effect is particularly pronounced for USB Insertion in clutter, where VLM guidance improves success from 10/20 to 15/20. The assistance of VLM reduces the difficulty of the task.

During preliminary model selection, we have evaluated several perception and vision-language models, including lightweight models such as YOLOE and SAM3 \cite{carion2026sam}, as well as larger models such as Qwen3-VL \cite{bai2025qwen3}, and the GPT-5.5 API \cite{singh2025openai}. Table 2 illustrates comparison of different VLM backbones on the charger insertion task. Although GPT-5.5 provided accurate predictions, its inference latency was too high for practical deployment. YOLOE and SAM3 produced results quickly, but were easily distracted by nearby objects with similar colors and struggled to localize targets accurately from semantic instructions. Qwen3-VL was less reliable than MolmoPoint on downstream tasks such as affordance prediction and point-based localization. They also occasionally produced large localization errors that could pose safety risks during robot execution. Considering the trade-off between inference speed and localization accuracy, we selected MolmoPoint.





\begin{table}[t]
    \centering
    \caption{Comparison of perception and VLM backbones on the charger insertion task. Inference time denotes the time required to localize the target in each episode. Success rate is evaluated over 10 real-robot trials.}
    \label{tab:vlm_comparison}
    \renewcommand{\arraystretch}{1.15}
    \setlength{\tabcolsep}{8pt}
    \begin{tabular}{lcc}
        \toprule
        \textbf{Perception/VLM Backbone}
        & \textbf{Inference Time (s) $\downarrow$}
        & \textbf{Success Rate $\uparrow$} \\
        \midrule
        YOLOE         & \textbf{0.07} & 0/10 (0\%) \\
        SAM 3         & 0.12          & 3/10 (30\%) \\
        Qwen3-VL-8B   & 2.00          & 5/10 (50\%) \\
        GPT-5.5 API   & 15.00         & \textbf{10/10 (100\%)} \\
        MolmoPoint-8B & 2.00          & \textbf{10/10 (100\%)} \\
        \bottomrule
    \end{tabular}
    
\end{table}

\subsection{Explore the scalability boundaries of S2C.}

To understand the generalization boundary of S2C, we conduct a more rigorous evaluation of its generalization and robustness across the following settings:
(1) Task A: variations in objects around the workspace.
(2) Task B: tablecloths of different colors.
(3) Task C: out-of-distribution objects appearing within the white power strip.
(4) Task D: strong light exposure, static or flashing.
(5) Task E: exposure to color light.
(6) Task F: slight external forces applied to the robotic arm.
(7) Task G: indoor lights are turned on or off.
(8) Task H: slight adjustments to the insertion position within a small range during execution.
(9) Task I: replacement with a black charger. 

As shown in Table~\ref{tab:generalization_boundary_transposed}, S2C exhibits strong robustness to Task A, B, D, E, F, H, G. However, the evaluation also reveals clear failure boundaries. S2C performs poorly when out-of-distribution objects appear inside the white power strip, especially for the Charger task, where the success rate drops to 0/10. Another important boundary appears when replacing the charger with a black charger. Both Charger and USB-Easy fail in this setting, indicating that S2C still struggles with severe object-level appearance shifts. Generally, these results illustrate that S2C generalizes well to environmental variations, including clutter in the workspace, changes in tablecloths, changes in illumination, slight physical disturbances, and small positional changes. Its main limitations arise from perturbations that change the appearance and geometry of the task-critical contact region.

\begin{table}[htbp]
\centering
\vspace{-0.1in}
\caption{Scalability evaluation of S2C across different tasks. Each task is tested 10 times.}
\label{tab:generalization_boundary_transposed}
\small
\begin{tabular}{lcccccccccc}
\toprule
 Task& A & B & C & D & E & F & G & H & I\\
\midrule
Charger & 10/10 & 9/10 & 0/10 & 10/10 & 8/10 & 10/10 & 10/10 & 10/10 & 0/10\\
USB-Easy & 10/10 & 8/10 & 3/10 & 10/10 & 8/10 & 10/10 & 10/10 & 10/10 & 0/10\\
\bottomrule
\end{tabular}

\end{table}

\begin{wraptable}{r}{0.7\textwidth}
  \centering
  \caption{Performance of S2C on additional contact-rich removal tasks}
  \label{tab:s2c-removal}
  \begin{tabular}{lcc}
    \toprule
    Task & Training Time (min) & Success Rate (\%) \\
    \midrule
    Test-tube Removal & 25 & 90 (9/10) \\
    Hard-drive Removal & 35 & 100 (10/10) \\
    \bottomrule
  \end{tabular}
\end{wraptable}

To add the task diversity and generalization, We have extended the application of S2C to some removal settings. To evaluate whether S2C generalizes beyond insertion tasks, we further tested the framework on two removal-oriented manipulation tasks: (1) grasping and extracting a test tube from a tube slot, and (2) grasping and removing a hard drive from a drive slot. These tasks require the robot to establish a stable grasp and execute a constrained extraction motion, rather than inserting an object into a target receptacle. In this experiment, we choose diffusion policy \cite{chi2025diffusion} as the base policy. We conducted 10 independent trials for each task in Table \ref{tab:s2c-removal}, and S2C successfully completed all trials, achieving a success rate of 100\% and 90\% on both tasks. These results provide additional evidence that the proposed semantic-to-contact decomposition is applicable not only to insertion, but also to contact-rich removal tasks. 

\vspace{-0.1in}
\section{Conclusion}
\vspace{-0.1in}
We present \textbf{Semantics-to-Contact}, a stagewise framework for robust contact-rich manipulation. S2C decomposes contact-rich manipulation into two stages: semantic focusing and contact refinement, where VLM-guided motion planning first brings the robot to a task-relevant pre-contact region, and Object-centric visual augmentation with human-in-the-loop real-world residual RL then enables visually robust, force-aware contact control. Across six real-world contact-rich tasks, S2C achieves strong robustness under positional shifts and diverse visual disturbances, demonstrating the effectiveness of integrating open-vocabulary perception with force-aware real-world learning. Future work will focus on extending S2C to robust long-horizon manipulation tasks.

\newpage
\section*{Acknowledgments} This work is supported in part by NSFC under Grant 62606447,
Shenzhen Science and Technology Program under grant KJZD20240903104008012, Shenzhen Science and Technology Program under grant ZDCY20250901113000001, CUHK-CUHK(SZ)-GDSTC Joint Collaboration Fund No. 2025A0505000053, 
and GuangDong Key Laboratory of Big Data Computing (2021B1212040002).

\bibliography{main}
\bibliographystyle{plainnat}
\newpage
\appendix

\section*{Supplementary Materials}

\section{Task Setup and Policy Training Details}
\label{appendix: task_setup}
In this section, we provide details regarding how each task is set up, including hardware and software; as
well as details on policy training.

\subsection{Tube Insertion (easy)}
In the Tube Insertion task, the robot is required to insert a cylindrical test tube into a fixed slot on a test-tube rack. The test tube is initially grasped by the robot and positioned above the rack with small variations in position and orientation. Successful completion of the task requires the robot to align the tube with the target slot and carefully guide it downward until it is fully inserted.

\subsubsection{Cropped Images}
\begin{figure}[h]
\centering
\begin{subfigure}[t]{0.23\linewidth}
  \centering
  \includegraphics[width=\linewidth]{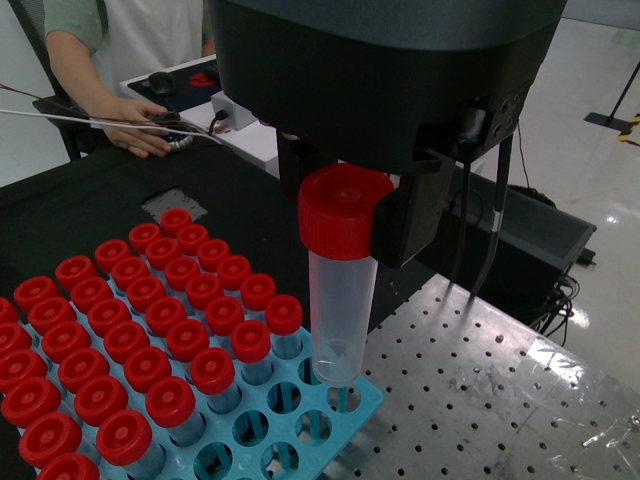}
  \caption*{wrist\_1}
\end{subfigure}
\hspace{1.2em}
\begin{subfigure}[t]{0.23\linewidth}
  \centering
  \includegraphics[width=\linewidth]{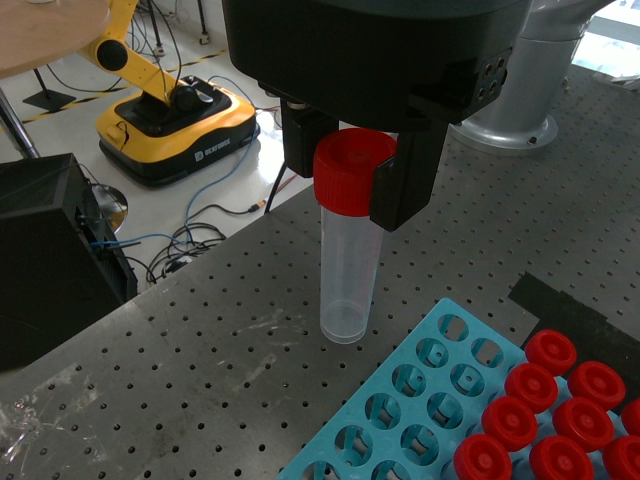}
  \caption*{wrist\_2}
\end{subfigure}
\caption{Sample input images from cameras used as inputs to the policy.}
\label{fig:cropped_card_insertion}
\end{figure}

We cropped the images to focus on the task-relevant parts of the scene, as shown in Fig.~\ref{fig:cropped_card_insertion}.

\subsubsection{Policy Training Details}
In Table~\ref{tab:card_train}, we report additional details of the policy training for this task.

\begin{table}[h]
\centering
\small
\caption{Policy training details for the Tube Insertion (easy) task.}
\begin{tabular}{l l}
\toprule
\textbf{Parameter} & \textbf{Value} \\
\midrule
base policy & ACT \\
Observation space & wrist\_1, wrist\_2, tcp\_pose, tcp\_vel, tcp\_f/t \\
Action space & 6D twist + 1D gripper \\
Classifier views & wrist\_1, wrist\_2 \\
Classifier accuracy & 99\% \\
Initial offline demonstrations & 30 \\
Max episode length & 100 environment steps \\
Reset method & Auto reset \\
Randomization range & 5 cm in x and y, 10 deg in rz \\
Proprio encoder size & 128 \\
Residual Policy MLP size & 256$\times$256 \\
Discount factor & 0.98 \\
Optimizer & Adam \\
xyz bound & 45cm*20cm*20cm\\
rpy bound & 0*0*$[-10, 10]deg$ \\
\bottomrule
\end{tabular}
\label{tab:card_train}
\end{table}

\subsection{Tube Insertion (medium)}
In the Sequential Tube Insertion task, the robot is required to insert a row of test tubes into multiple slots on a test-tube rack in a predefined order. Each test tube is grasped by the robot and must be aligned with the next target slot before insertion. Successful completion of the task requires the robot to repeatedly localize the correct empty slot, adjust the tube pose, and perform stable downward insertion without disturbing previously placed tubes.

\subsubsection{Cropped Images}
\begin{figure}[h]
\centering
\begin{subfigure}[t]{0.23\linewidth}
  \centering
  \includegraphics[width=\linewidth]{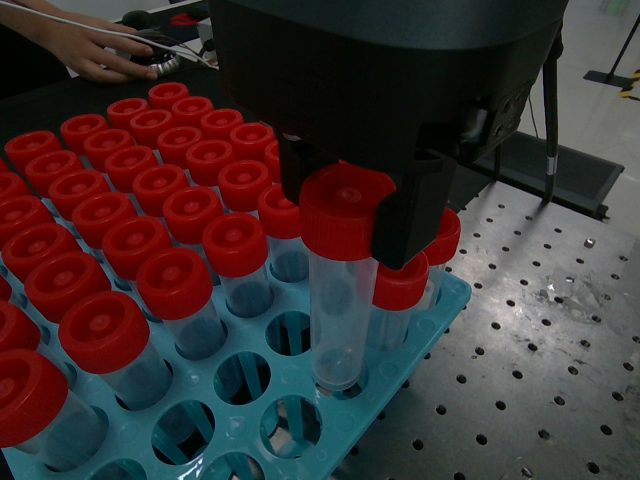}
  \caption*{wrist\_1}
\end{subfigure}
\hspace{1.2em}
\begin{subfigure}[t]{0.23\linewidth}
  \centering
  \includegraphics[width=\linewidth]{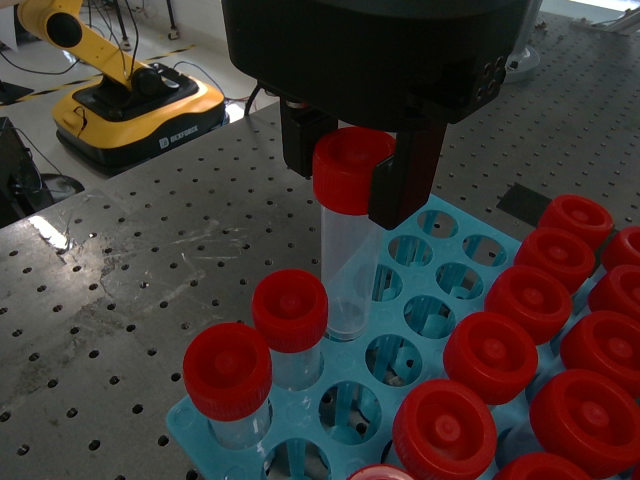}
  \caption*{wrist\_2}
\end{subfigure}
\caption{Sample input images from cameras used as inputs to the policy.}
\label{fig:cropped_door_opening}
\end{figure}

Fig.~\ref{fig:cropped_door_opening} shows the hardware setup for the motherboard assembly task, which presents the robot, the camera
placements, and the task arrangement.

\subsubsection{Policy Training Details}
In Table~\ref{tab:door_train}, we report additional details of the policy training for this task.

\begin{table}[h]
\centering
\small
\caption{Policy training details for the Tube Insertion (medium) task.}
\begin{tabular}{l l}
\toprule
\textbf{Parameter} & \textbf{Value} \\
\midrule
base policy & ACT \\
Observation space & wrist\_1, wrist\_2, tcp\_pose, tcp\_vel, tcp\_f/t \\
Action space & 6D twist+ 1D gripper \\
Classifier views & wrist\_1, wrist\_2 \\
Classifier accuracy & 99\% \\
Initial offline demonstrations & 40 \\
Max episode length & 100 environment steps \\
Reset method & Auto reset \\
Randomization range & 5 cm in x and y, 10 deg in rz \\
Proprio encoder size & 128 \\
Residual Policy MLP size & 256$\times$256 \\
Discount factor & 0.98 \\
Optimizer & Adam \\
xyz bound & 45cm*30cm*30cm\\
rpy bound & 0*0*$[-10, 10]deg$ \\
\bottomrule
\end{tabular}
\label{tab:door_train}
\end{table}

\subsection{Charger grasp-insertion}
In the Charger Grasp-Insertion task, the robot first grasps a charging plug and then precisely inserts it into a corresponding socket on a power strip. The plug is initially placed at a nearby location with small variations in pose. Successful completion of the task requires accurate grasping, reliable pose alignment between the plug and the socket, and careful control during the insertion phase to avoid misalignment or collisions. This task emphasizes sequential manipulation, fine-grained spatial alignment, and contact-aware insertion under tight geometric constraints.

\subsubsection{Cropped Images}
\begin{figure}[h]
\centering
\begin{subfigure}[t]{0.23\linewidth}
  \centering
  \includegraphics[width=\linewidth]{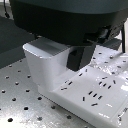}
  \caption*{wrist\_1}
\end{subfigure}
\hspace{1.2em}
\begin{subfigure}[t]{0.23\linewidth}
  \centering
  \includegraphics[width=\linewidth]{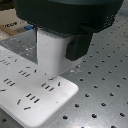}
  \caption*{wrist\_2}
\end{subfigure}
\caption{Sample input images from cameras used as inputs to the policy.}
\label{fig:cropped_Charger_grasp-insertion}
\end{figure}

Fig.~\ref{fig:cropped_Charger_grasp-insertion} shows the hardware setup for the motherboard assembly task, which presents the robot, the camera
placements, and the task arrangement.

\subsubsection{Policy Training Details}
In Table~\ref{tab:door_train}, we report additional details of the policy training for this task.

\begin{table}[h]
\centering
\small
\caption{Policy training details for the Charger grasp-insertion.}
\begin{tabular}{l l}
\toprule
\textbf{Parameter} & \textbf{Value} \\
\midrule
base policy & ACT \\
Observation space & wrist\_1, wrist\_2, tcp\_pose, tcp\_vel, tcp\_f/t \\
Action space & 6D twist + 1D gripper \\
Classifier views & wrist\_1, wrist\_2 \\
Classifier accuracy & 99\% \\
Initial offline demonstrations & 38 \\
Max episode length & 100 environment steps \\
Reset method & Auto reset \\
Randomization range & 5 cm in x and y, 30 deg in rz \\
Proprio encoder size & 128 \\
Residual Policy MLP size & 256$\times$256 \\
Discount factor & 0.98 \\
Optimizer & Adam \\
xyz bound & 45cm*30cm*30cm\\
rpy bound & 0*0*$[-60, 60]deg$ \\
\bottomrule
\end{tabular}
\label{tab:hang_train}
\end{table}

\subsection{Tube Insertion (hard)}
In the Hard Tube Insertion task, the robot must insert a test tube into whichever slot is empty in a partially occupied test-tube rack. The empty slot varies across trials, requiring the robot to visually identify the available opening before performing insertion. Successful completion requires accurate slot localization, precise alignment, and stable contact-aware insertion while avoiding collisions with neighboring tubes and the rack.

\subsubsection{Cropped Images}
\begin{figure}[h]
\centering
\begin{subfigure}[t]{0.23\linewidth}
  \centering
  \includegraphics[width=\linewidth]{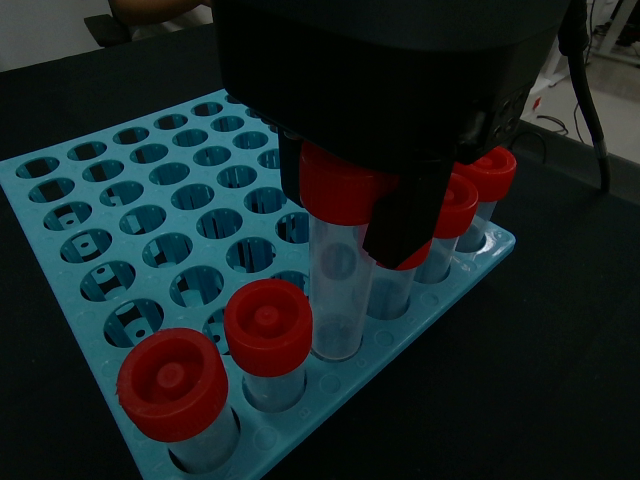}
  \caption*{wrist\_1}
\end{subfigure}
\hspace{1.2em}
\begin{subfigure}[t]{0.23\linewidth}
  \centering
  \includegraphics[width=\linewidth]{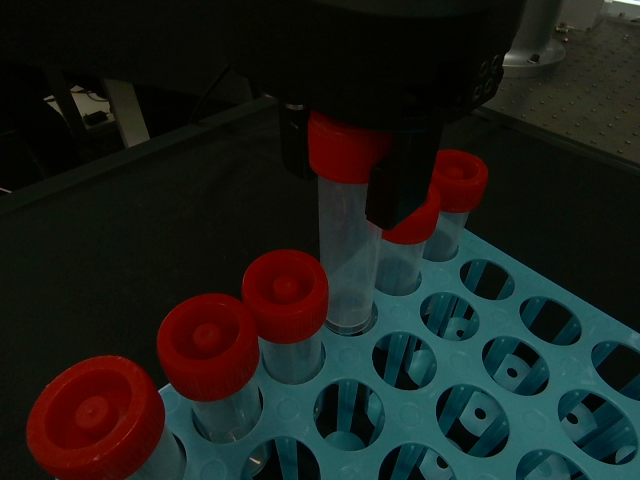}
  \caption*{wrist\_2}
\end{subfigure}
\caption{Sample input images from cameras used as inputs to the policy.}
\label{fig:cropped_Keychain_hanging}
\end{figure}

Fig.~\ref{fig:cropped_Keychain_hanging} shows the hardware setup for the motherboard assembly task, which presents the robot, the camera
placements, and the task arrangement.

\subsubsection{Policy Training Details}
In Table~\ref{tab:Keychain_hanging}, we report additional details of the policy training for this task.

\begin{table}[h]
\centering
\small
\caption{Policy training details for the Tube Insertion (hard) task.}
\begin{tabular}{l l}
\toprule
\textbf{Parameter} & \textbf{Value} \\
\midrule
base policy & ACT \\
Observation space & wrist\_1, wrist\_2, tcp\_pose, tcp\_vel, tcp\_f/t \\
Action space & 6D twist + 1D gripper \\
Classifier views & wrist\_1, wrist\_2 \\
Classifier accuracy & 99\% \\
Initial offline demonstrations & 40 \\
Max episode length & 100 environment steps \\
Reset method & Auto reset \\
Randomization range & 5 cm in x and y, 10 deg in rz \\
Proprio encoder size & 128 \\
Residual Policy MLP size & 256$\times$256 \\
Discount factor & 0.98 \\
Optimizer & Adam \\
xyz bound & 45cm*30cm*30cm\\
rpy bound & 0*0*$[-10, 10]deg$ \\
\bottomrule
\end{tabular}
\label{tab:Keychain_hanging}
\end{table}

\subsection{USB Insertion (easy)}
In the USB Insertion (Easy) task, the robot starts with a USB plug already held in the gripper and inserts it into a USB port on a charging adapter. The charging adapter is fixed in position and orientation throughout the task. Successful completion requires accurate alignment between the USB plug and the port and careful control during the insertion motion to avoid misalignment or collisions with the port edges. This variant primarily evaluates precise insertion control under constrained geometry with a known target pose.

\subsubsection{Cropped Images}
\begin{figure}[h]
\centering
\begin{subfigure}[t]{0.23\linewidth}
  \centering
  \includegraphics[width=\linewidth]{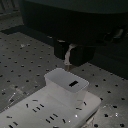}
  \caption*{wrist\_1}
\end{subfigure}
\hspace{1.2em}
\begin{subfigure}[t]{0.23\linewidth}
  \centering
  \includegraphics[width=\linewidth]{Figure/usb_wrist1.jpg}
  \caption*{wrist\_2}
\end{subfigure}
\caption{Sample input images from cameras used as inputs to the policy.}
\label{fig:cropped_USB_Insertion}
\end{figure}

Fig.~\ref{fig:cropped_USB_Insertion} shows the hardware setup for the motherboard assembly task, which presents the robot, the camera
placements, and the task arrangement.

\subsubsection{Policy Training Details}
In Table~\ref{tab:USB_Insertion}, we report additional details of the policy training for this task.

\begin{table}[h]
\centering
\small
\caption{Policy training details for the USB Insertion (easy) task.}
\begin{tabular}{l l}
\toprule
\textbf{Parameter} & \textbf{Value} \\
\midrule
base policy & ACT \\
Observation space & wrist\_1, wrist\_2, tcp\_pose, tcp\_vel, tcp\_f/t \\
Action space & 6D twist \\
Classifier views & wrist\_1, wrist\_2 \\
Classifier accuracy & 99\% \\
Initial offline demonstrations & 35 \\
Max episode length & 100 environment steps \\
Reset method & Auto reset \\
Randomization range & 5 cm in x and y, 30 deg in rz \\
Proprio encoder size & 128 \\
Residual Policy MLP size & 256$\times$256 \\
Discount factor & 0.98 \\
Optimizer & Adam \\
xyz bound & 50cm*25cm*25cm\\
rpy bound & 0*0*$[-60, 60]deg$ \\
\bottomrule
\end{tabular}
\label{tab:USB_Insertion}
\end{table}

\subsection{USB Insertion (hard)}
In the USB Insertion (Hard) task, the robot starts with a USB plug already held in the gripper and inserts it into a USB port on a charging adapter whose position is randomly selected from a set of predefined locations. The target port pose therefore varies across episodes. Successful completion requires the policy to first localize the target port and then perform accurate pose alignment and contact-aware insertion. Compared to the easy variant, this task introduces target pose uncertainty and emphasizes robustness to spatial variation in addition to precise insertion control.

\subsubsection{Cropped Images}
\begin{figure}[h]
\centering


\begin{subfigure}{0.48\linewidth}
    \centering
    \includegraphics[width=0.48\linewidth]{Figure/usb_wrist1.jpg}
    \hfill
    \includegraphics[width=0.48\linewidth]{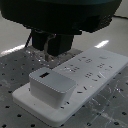}
    \caption*{Position 1}
\end{subfigure}
\hfill
\begin{subfigure}{0.48\linewidth}
    \centering
    \includegraphics[width=0.48\linewidth]{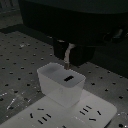}
    \hfill
    \includegraphics[width=0.48\linewidth]{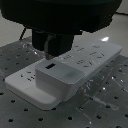}
    \caption*{Position 2}
\end{subfigure}

\vspace{0.5em}


\begin{subfigure}{0.48\linewidth}
    \centering
    \includegraphics[width=0.48\linewidth]{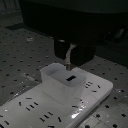}
    \hfill
    \includegraphics[width=0.48\linewidth]{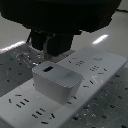}
    \caption*{Position 3}
\end{subfigure}
\hfill
\begin{subfigure}{0.48\linewidth}
    \centering
    \includegraphics[width=0.48\linewidth]{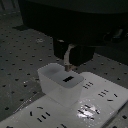}
    \hfill
    \includegraphics[width=0.48\linewidth]{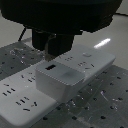}
    \caption*{Position 4}
\end{subfigure}

\caption{Sample input images from cameras used as inputs to the policy.}
\label{fig:cropped_USB_Insertion_hard}
\end{figure}

Fig.~\ref{fig:cropped_USB_Insertion_hard} shows the hardware setup for the motherboard assembly task, which presents the robot, the camera
placements, and the task arrangement.

\subsubsection{Policy Training Details}
In Table~\ref{tab:USB_Insertion_hard}, we report additional details of the policy training for this task.

\begin{table}[h]
\centering
\small
\caption{Policy training details for the USB Insertion (hard) task.}
\begin{tabular}{l l}
\toprule
\textbf{Parameter} & \textbf{Value} \\
\midrule
base policy & ACT \\
Observation space & wrist\_1, wrist\_2, tcp\_pose, tcp\_vel, tcp\_f/t \\
Action space & 6D twist \\
Classifier views & wrist\_1, wrist\_2 \\
Classifier accuracy & 99\% \\
Initial offline demonstrations & 35 \\
Max episode length & 100 environment steps \\
Reset method & Auto reset \\
Randomization range & 10 cm in x and y, 30 deg in rz \\
Proprio encoder size & 128 \\
Residual Policy MLP size & 256$\times$256 \\
Discount factor & 0.98 \\
Optimizer & Adam \\
xyz bound & 50cm*25cm*25cm\\
rpy bound & 0*0*$[-60, 60]deg$ \\
\bottomrule
\end{tabular}
\label{tab:USB_Insertion_hard}
\end{table}

\section{Running Setting}
\label{appendix: running_setting}

To ensure a fair comparison, we train Octo-small and ACT using 50 data samples. We train ACT for 10000 epochs and Octo-small for 20000 epochs. For SERL, HIL-ConRFT, HIL-SERL and S2C, we initially used 30-40 expert trajectories to fill $D_{offline}$
. The $\pi_{base}$ for S2C was pretrained with 40 data samples for 5000 epochs. SERL, HIL-ConRFT, and HIL-SERL all use BC to pretrain the actor and critic. SERL and HIL-SERL use an MLP network for BC, while HIL-ConRFT uses Octo-small. We deploy the VLM on A800 GPUs, communicating via client-server. The RWRL method was trained using 5090 GPUs, while the imitation learning method was trained using A800 GPUs.

\section{Visual Augmentation Details}
\label{appendix: Augmentation}
In practice, promptable segmentation may be imperfect when objects are transparent or visually similar to the background. To improve mask quality, we optionally calibrate the YOLOE visual features with a lightweight linear probe. Specifically, given a small set of labeled foreground masks $\mathcal{D}_{\mathrm{mask}}=\{(P_i,M_i^{*})\}$, we freeze the YOLOE backbone and train a linear segmentation head $h_{\psi}$:
\begin{equation}
    \psi^{*}
    =
    \arg\min_{\psi}
    \mathbb{E}_{(P_i,M_i^{*})\sim \mathcal{D}_{\mathrm{mask}}}
    \left[
    \mathcal{L}_{\mathrm{seg}}
    \left(
    h_{\psi}
    \left(
    f_{\mathrm{YOLOE}}(P_i)
    \right),
    M_i^{*}
    \right)
    \right],
    \label{eq:linear_probe}
\end{equation}

\begin{figure}[t]
    \centering
    \includegraphics[width=0.99\linewidth]{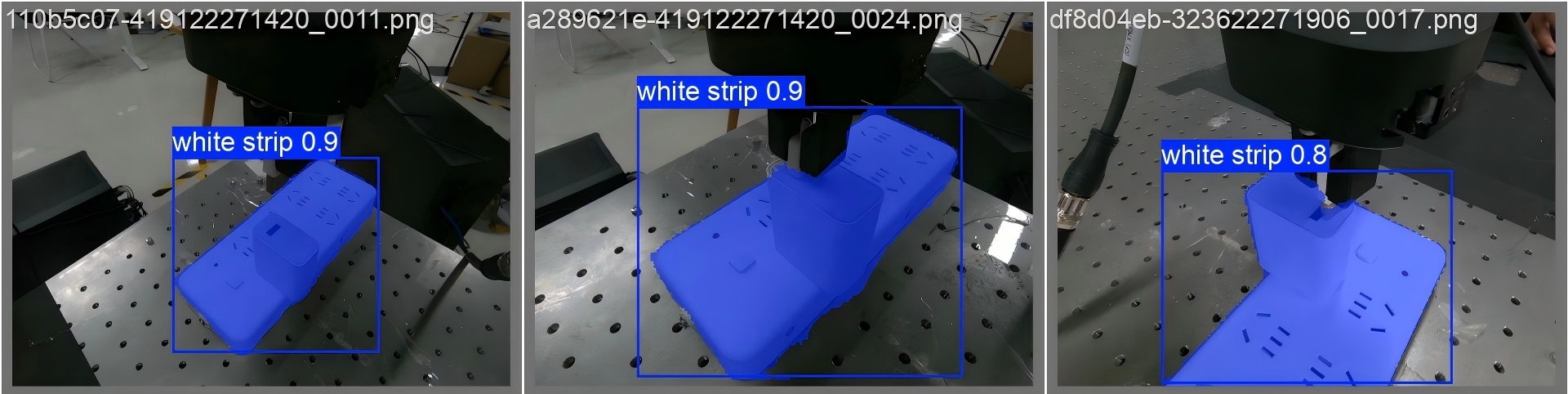}
    \caption{YOLOE segmentation validation examples.}
    \label{fig:single_image}
\end{figure}

We use the polygon tool in Label Studio \cite{Label_Studio} to annotate the categories to be fine-tuned. Figure \ref{fig:single_image} shows examples of the results on the validation set after fine-tuning.

\section{VLM instructions}
\label{sec: vlm_instructions}
USB (Easy / Hard): “Point to the square USB port on the white charger.”
Charger: “Point to the center of the lower-left three-hole socket on the white power strip placed on the tablecloth.”
Tube (Easy / Medium): “Point to the first empty slot from the top in the rightmost column of the blue test-tube rack.”
Tube (Hard): “Point to the red test tube in the middle of the rightmost column of the blue test-tube rack.”

\newpage
\section*{NeurIPS Paper Checklist}

\begin{enumerate}

\item {\bf Claims}
    \item[] Question: Do the main claims made in the abstract and introduction accurately reflect the paper's contributions and scope?
    \item[] Answer: \answerYes{} 
    \item[] Justification: We have introduced the main claim in our abstract and introduction.

\item {\bf Limitations}
    \item[] Question: Does the paper discuss the limitations of the work performed by the authors?
    \item[] Answer:\answerYes{} 
    \item[] Justification: We discussed our limitations in Section 5.3.

\item {\bf Theory assumptions and proofs}
    \item[] Question: For each theoretical result, does the paper provide the full set of assumptions and a complete (and correct) proof?
    \item[] Answer: \answerNA{} 
    \item[] Justification: This paper is an application-oriented piece about robots, and does not involve theoretical proofs or assumptions.

    \item {\bf Experimental result reproducibility}
    \item[] Question: Does the paper fully disclose all the information needed to reproduce the main experimental results of the paper to the extent that it affects the main claims and/or conclusions of the paper (regardless of whether the code and data are provided or not)?
    \item[] Answer: \answerYes{} 
    \item[] Justification: We have fully described the methods and experimental details in Sections 4, 5, and the supplementary materials. At the same time, we have many videos demonstrating the effectiveness of our method.

\item {\bf Open access to data and code}
    \item[] Question: Does the paper provide open access to the data and code, with sufficient instructions to faithfully reproduce the main experimental results, as described in supplemental material?
    \item[] Answer: \answerNo{} 
    \item[] Justification: After acceptance, we will provide the code and data in the future.

\item {\bf Experimental setting/details}
    \item[] Question: Does the paper specify all the training and test details (e.g., data splits, hyperparameters, how they were chosen, type of optimizer) necessary to understand the results?
    \item[] Answer:\answerYes{} 
    \item[] Justification: We have fully described experimental details in Sections 5 and the supplementary materials.


\item {\bf Experiment statistical significance}
    \item[] Question: Does the paper report error bars suitably and correctly defined or other appropriate information about the statistical significance of the experiments?
    \item[] Answer: \answerYes{} 
    \item[] Justification: For each task, we have evaluated multiple times under different settings and calculated the average success rate metric before calculating the success rate.

\item {\bf Experiments compute resources}
    \item[] Question: For each experiment, does the paper provide sufficient information on the computer resources (type of compute workers, memory, time of execution) needed to reproduce the experiments?
    \item[] Answer: \answerYes{} 
    \item[] Justification: 
    We have fully described the computer resources in the supplementary materials.
    
\item {\bf Code of ethics}
    \item[] Question: Does the research conducted in the paper conform, in every respect, with the NeurIPS Code of Ethics \url{https://neurips.cc/public/EthicsGuidelines}?
    \item[] Answer: \answerYes{} 
    \item[] Justification: It is satisfied with \url{https://neurips.cc/public/EthicsGuidelines}.

\item {\bf Broader impacts}
    \item[] Question: Does the paper discuss both potential positive societal impacts and negative societal impacts of the work performed?
    \item[] Answer: \answerNA{} 
    \item[] Justification: There is no societal impact of the work performed.

    
\item {\bf Safeguards}
    \item[] Question: Does the paper describe safeguards that have been put in place for responsible release of data or models that have a high risk for misuse (e.g., pre-trained language models, image generators, or scraped datasets)?
    \item[] Answer: \answerNA{} 
    \item[] Justification: The paper is not related to a high-risk model or data.

\item {\bf Licenses for existing assets}
    \item[] Question: Are the creators or original owners of assets (e.g., code, data, models), used in the paper, properly credited and are the license and terms of use explicitly mentioned and properly respected?
    \item[] Answer: \answerYes{} 
    \item[] Justification: We have provided proper citations for all of them.

\item {\bf New assets}
    \item[] Question: Are new assets introduced in the paper well documented and is the documentation provided alongside the assets?
    \item[] Answer: \answerNA{} 
    \item[] Justification: The paper does not introduce a new asset, dataset or benchmark.

\item {\bf Crowdsourcing and research with human subjects}
    \item[] Question: For crowdsourcing experiments and research with human subjects, does the paper include the full text of instructions given to participants and screenshots, if applicable, as well as details about compensation (if any)? 
    \item[] Answer: \answerNA{} 
    \item[] Justification: Our experiments are not related to human subjects.

\item {\bf Institutional review board (IRB) approvals or equivalent for research with human subjects}
    \item[] Question: Does the paper describe potential risks incurred by study participants, whether such risks were disclosed to the subjects, and whether Institutional Review Board (IRB) approvals (or an equivalent approval/review based on the requirements of your country or institution) were obtained?
    \item[] Answer: \answerNA{} 
    \item[] Justification: Our experiments are not related to human subjects.

\item {\bf Declaration of LLM usage}
    \item[] Question: Does the paper describe the usage of LLMs if it is an important, original, or non-standard component of the core methods in this research? Note that if the LLM is used only for writing, editing, or formatting purposes and does \emph{not} impact the core methodology, scientific rigor, or originality of the research, declaration is not required.
    \item[] Answer: \answerNA{} 
    \item[] Justification: The LLM is used only for writing, editing, or formatting purposes

\end{enumerate}

\end{document}